\documentclass[11pt,a4,english]{article}
\usepackage[latin9]{inputenc}
\usepackage{geometry}
\usepackage{natbib}
\usepackage{verbatim}
\usepackage{amsmath}
\usepackage{graphicx}
\usepackage{longtable}
\usepackage{setspace}
\usepackage{amsfonts}
\usepackage{lscape}
\usepackage[colorlinks=true,linkcolor=blue, citecolor=blue]{hyperref}

\makeatletter
\usepackage{graphicx}
\makeatother
\usepackage{babel}
\usepackage{amsfonts}
\usepackage{amsthm}
\usepackage[toc,page]{appendix}
\usepackage{booktabs,caption}
\usepackage{multirow}
\usepackage[labelfont=bf,textfont=md]{caption}
\usepackage[labelsep=space]{caption}
\usepackage{hyperref}
\usepackage{lmodern}
\usepackage[T1]{fontenc}
\usepackage{mathtools}

\makeatletter
\newcommand{\nextverbatimspread}[1]{%
  \def\verbatim@font{%
    \linespread{#1}\normalfont\ttfamily
    \gdef\verbatim@font{\normalfont\ttfamily}}
}
\makeatother

\usepackage{natbib}
\usepackage{bibentry}
\nobibliography*

\theoremstyle{definition}
\newtheorem{theorem}{Theorem}

\newtheorem{assumption}{Assumption}

\newtheorem{example}{Example}

\newtheorem{remark}{Remark}

\usepackage{amsmath}
\newcommand\norm[1]{\left\lVert#1\right\rVert}

\begin{document}
\pagenumbering{roman}

\title{ {\Large \textbf{Forecast Evaluation in Large Cross-Sections of Realized Volatility}\thanks{The current manuscript was prepared at the Department of Economics of the University of Southampton. I would like to thank Jose Olmo, Jean-Yves Pitarakis and  Tassos Magdalinos for guidance and continuous encouragement throughout the PhD programme. Financial support from the VC PhD Scholarship of the University of Southampton is gratefully acknowledged. Furthermore, the author acknowledge the use of Iridis4 HPC Facility and associated support services at the University of Southampton in the completion of this work. } \\
}
}

\author{
\textbf{Christis Katsouris}\thanks{Ph.D. Candidate, Department of Economics, University of Southampton, Southampton, SO17 1BJ, UK. \textit{E-mail}: \textcolor{blue}{C.Katsouris@soton.ac.uk}.}
}

\date{September 28, 2020}

\maketitle

\begin{abstract}
\vspace*{-0.5 em}
In this paper, we consider the forecast evaluation of realized volatility measures under cross-section dependence using equal predictive accuracy testing procedures. We evaluate the predictive accuracy of the model based on the augmented cross-section when forecasting Realized Volatility. Under the null hypothesis of equal predictive accuracy the benchmark model employed is a standard HAR model while under the alternative of non-equal predictive accuracy the forecast model is an augmented HAR model estimated via the LASSO shrinkage. We study the sensitivity of forecasts to the model specification by incorporating a measurement error correction as well as cross-sectional jump component measures. The out-of-sample forecast evaluation of the models is assessed with numerical implementations.
\\

\textbf{Keywords:} equal predictive accuracy testing, realized volatility measures, correlated covariates, lasso shrinkage, penalty functions.   
\end{abstract}

\newpage 

\setcounter{page}{1}
\pagenumbering{arabic}

\section{Introduction}

Forecasting volatility has a fundamental scope for financial economics with applications in asset pricing, risk management as well as systemic risk monitoring due to the fact that forecasts of asset return volatilities are essential inputs for pricing models (\cite{bollerslev2020realized}). A vast body of literature has been devoted to model design capable of accurately capturing volatility dynamics and producing reliable volatility forecasts. Furthermore, the increasing availability of high frequency data pushed the development of methods such as latent variable models such as the GARCH specifications as well as models for Stochastic Volatility (as in \cite{bollerslev1986generalized} and \cite{hansen2005forecast}). Moreover, the inclusion of high frequency filters via the use of estimators for the true latent integrated volatilities has been examined in various studies such as in \cite{andersen1998answering}, \cite{barndorff2002econometric}, \cite{andersen2001distribution}, \cite{andersen2003modeling}, \cite{andersen2007roughing} and \cite{ait2014high}. 

In practise, time series observations for realized volatility measures at a given frequency (such as daily) are typically obtained by summing higher frequency squared returns (e.g. intra-day) and can therefore be readily constructed in a model-free way from high frequency asset pricing data. This approach essentially turns unobserved market volatility into an observable time series which can be modelled via standard time series techniques. In this framework, a functional form that has been found to be particularly successful in its ability to capture the stylised facts of volatility (e.g. long-memory and persistence) is the heterogeneous autoregressive (HAR) specification that models RV series as multi-component autoregressions in which each component captures volatility at different time scales (see, \cite{corsi2009simple}). As hinted by the autoregressive structure of this class of models, HAR based RV forecasts of a given asset are solely based on the information contained in its own past volatilities, potentially missing out on valuable information that may be provided by other assets' volatilities. According to \cite{diebold2014network}, an important stylised fact characterising financial markets is their strong interdependence which is itself known to be increasing during periods of uncertainty (see, also \cite{billio2012econometric}). 

The main goal of this paper is to introduce a novel augmented HAR type specification that is explicitly designed to extend the information set for second-order moments under cross-sectional dependence. In terms of the empirical application, this is equivalent to incorporating the volatility dynamics of the cross-section in addition to lagged volatility series of firm $j$. The particular modelling approach motivates the proposition of a cross-sectionally augmented specification for improving the out-of-sample predictive accuracy of standard HAR models. Although one may conjecture that assets in similar industries or assets that are subject to common economic shocks are candidate predictors of an assets own future RV, the breadth of possible predictor choices and the typical size of the relevant cross-section  requires us to consider shrinkage based estimation techniques that are designed to handle such data rich environments. Our paper contributes to the literature of cross-sectional predictability of second-order stock return moments based on high frequency filters such as the realized volatility measures.  

A notable novelty of our econometric framework is the focus on a high dimensional setting that incorporates information on past RVs from the entire cross-section of assets of interest. More specifically, as we wish to remain agnostic about the choice of predictors and let the data decide, our augmented HAR specification is estimated via LASSO based techniques as proposed by \cite{tibshirani1996regression} that can simultaneously perform variable selection and regularisation. Furthermore, as our objective is to assess the relevance of cross-sectional information via its potential out-of-sample contributions to improving individual forecasts, the Lasso and its variants are implemented in a pseudo out-of-sample fashion by considering both recursive and rolling window based forecasting schemes. In particular, the constructed out-of-sample forecast error sequences allow us to implement formal predictive accuracy tests a la \cite{diebold2002comparing}, \cite{clark2007approximately} and \cite{giacomini2006tests}. The pool of competing models that we consider also extends beyond the two econometric specifications mentioned above due to the need to accommodate microstructure noise, jumps and measurement errors.

\subsection{Literature Review}

Forecasting Realized Volatility via the Cross-Section is indeed not so common in the financial econometrics literature even though the use of aggregate market risks determined by large Cross-Sections is a widely used idea (see, e.g., \cite{bollerslev2018risk}, \cite{doshi2019leverage} and \cite{li2017jump}). Our proposed econometric framework is based on statistical estimation of "pooled risks" as well as variable selection via the Lasso norm. In particular, the Lasso penalization and variable selection has been recently introduced for the cross-sectional identification of candidate predictors for return predictability. A related methodological approach applied towards this direction is presented by \cite{audrino2016lassoing}. Our work is also closely related to the approach of \cite{chinco2019sparse}. More specifically, the authors focus on the identification of rare sparse predictors within particular intervals in order to examine the predictive accuracy of a Cross-Section of lagged returns for rolling one-minute-ahead return forecasts using the Lasso shrinkage methodology. Further applications similar to our study include the paper of \cite{yao2019novel} in which the authors introduce a group cluster Lasso specification for the Augmented HAR model.

Our paper, covers within a unified framework important aspects of forecasting second-order moments of stock returns using Large Cross-Sections. Specifically, the use of large cross-sections for forecasting RV provides statistical validity of predictability via the cross-section. Investigating the forecasting ability of the cross-section set has been previously discussed by \cite{white2000reality}, who mentions: \textit{"one can test whether a neural network of apparently optimal complexity provides a true improvement over a simple benchmark, e.g., a zero hidden unit model"}. In this paper, we represent the aforementioned forecasting framework by comparing the forecast performance of a benchmark model with a set of regressors the lagged RVs of the firm under examination, and the forecast performance of the forecast model which incorporates as regressors the lagged realized volatility measures of the cross-section.  

Firstly, the cross-section of stock returns is examined by the seminal paper \cite{eugene1992cross}. Furthermore, due to the vast availability of high frequency data further applications within this settings include the examination of predictive accuracy in models for such high frequency data (see, for example \cite{galvao2013changes}). Additionally, other frameworks examine how fluctuations in the micro level affects aggregate macroeconomic behaviour such as the study of  \cite{gabaix2011granular}. While more recently, the literature focuses on the examination of the specific aspects of high dimensional models, such as the assumption of sparsity affects the predictive ability (see, for example \cite{giannone2017economic}). The pooling of cross-section and time series data is a topic which has been extensively examined in the time series econometrics literature. More specifically, related literature to the asymptotic theory of cross-sectional estimators can be found in  \cite{balestra1966pooling}, \cite{wallace1969use}, \cite{maddala1971use}, \cite{mundlak1978pooling} and \cite{andrews2005cross} among others. The particular methodologies are concerned with estimation robustness and consistency using both the cross-section and time series dimension. Therefore, we identify a gap in the literature which investigate the forecasting accuracy when cross-sectional model specifications with mixed frequency data (realized volatility filters) are utilized within a predictive accuracy environment. 

Secondly, nonlinear panel data models have been proposed to examine cross-sectional dependence such as in the paper of \cite{kapetanios2014nonlinear}. In particular, regression models with mixed sampling frequencies have been developed for the purpose of providing parsimonious representations when the dependent variable is of different sampling frequency than the regressors (see, \cite{andreou2010regression} and \cite{andreou2016use}).  Furthermore, nonlinear regression models (NLS) have been proposed in the literature as suitable specifications for cross-sectional model specifications using mixed frequency data. A recent framework for forecasting using panel data is presented by \cite{liu2020forecasting} who give emphasis on the cross-sectional distribution of statistics rather than the encompassing property of the baseline model versus the forecast model, as in the high dimensional environment examined in this paper. The authors, using a random effect panel data specification derive some interesting results regarding the estimated expected loss or risk when forecasting using panel data estimators. Moreover, the study of \cite{andreou2019inference} propose a unified framework which encompasses both the time series and cross-section dimension specifically for mixed frequency data. 

In terms of the cross-sectional predictability of stock returns a notable framework is the recent study of \cite{chinco2019sparse} who examine the aspects of episodic predictive ability of predictors as sparse signals from the cross-section. Specifically, to identify rare sparse predictors (e.g., predictors which are sufficiently unexpected and short-lived) within particular intervals, the authors consider the predictive accuracy of a cross-section of lagged returns for rolling one-minute-ahead return forecasts using shrinkage methodologies. The Lasso shrinkage estimator appears to increase both out-of-sample fit and forecast-implied Sharpe ratios. Moreover, the selection of predictors even though appears to be sparse and solely based on the statistical properties of the Lasso shrinkage in identifying a set of predictors, it appears to be related to certain economic events and for stocks with certain characteristics. This particular feature can be employed in portfolio allocation problems, incorporating the cross-section information set in the optimal portfolio choice of investors under the assumptions of a mean-variance framework. A related recent study is presented by \cite{mcgee2020optimal}. 

The problem of econometric determination of a model with a data-based approach it has been previously discussed in the literature. \cite{phillips1996econometric} studies various econometric methodologies for the model determination based on the given sample of observations. The particular approach emphasizes the fact that using a penalize likelihood methodology is justified even if one does not take the Bayesian approach. In this paper, we investigate the properties of high-dimensional variable selection methods for correlated data using realized volatility measures based on a cross-section of firms. The paper is organized as follows. Section 2, introduces the model for estimating and forecasting  Realized Volatility. Section 3 presents the forecast evaluation methodologies and implements the out-of-sample comparisons for both nested and non-nested HAR environments for on a large RV dataset. Section 4, concludes. 

\newpage

\section{High-Dimensional Cross-Section Regression Model}

The proposed statistical framework constitutes, to the best of our knowledge, the first attempt in bridging the gap in the literature of modelling environments for higher order moments of stock returns under cross-sectional dependence. Thus, we consider predictive accuracy testing procedures based on these high-dimensional cross-section time series regression models. Furthermore, classical estimation methods such as the OLS estimator is found to perform poorly in high-dimensional estimation environments under multicollinearity which motivates the implementation of the Lasso shrinkage norm.

\subsection{Model and Assumptions}
\label{Section2.1}

Consider the cross-section predictive regression model for an $h-$period forecast horizon
\begin{align}
\label{specification}
y_{i,t+h} = \phi_{i0} + \sum_{j=1}^N  \text{\boldmath$\phi$}_{ij}^{\prime} \mathbf{X}_{j,t}  + \epsilon_{i, t+h}, \ \ \ t = 1,...,T. 
\end{align}
where  $y_{i,t+h}$ is a scalar dependent variable, $X_{i,t}$ is a $p-$dimensional vector of predictors, where $p \equiv p_N$, a function of the cross-section units. Denote with $\epsilon_{i, t}$ the unobserved scalar noise such that $\mathbb{E} \left( \epsilon_{i, t} | X_t \right)=0$ and $\text{\boldmath$\phi$}_{i}$ the unknown parameter vector, where $\text{\boldmath$\phi$}_{i} = ( \phi_{i,0}, \phi_{i,1}, \phi_{i,2},...., \phi_{i,N} )^{\prime}$ is estimated by minimizing the following objective function
\begin{align}
 Q_T ( \text{\boldmath$\phi$}_{i} )  = \left\{ \frac{1}{T} \sum_{t=1}^T \left(  y_{i,t+h} - \phi_{i,0} - \sum_{j=1}^N \text{\boldmath$\phi$}_{ij}^{\prime} \mathbf{X}_{j,t} \right)^2 \right\}
\end{align}
When $N$ is large we assume a sparse structure for the parameter vector $\text{\boldmath$\phi$}_{i}$ and thus consider a penalised estimation approach whereby $\hat{\text{\boldmath$\phi$}_{i}}  =  \underset{ \text{\boldmath$\phi$} \in \mathbb{R}^{N+1} }{  \text{argmin} }  \ Q_T ( \text{\boldmath$\phi$}_{i} )$, which implies that  
\begin{align}
\label{estimation}
\widehat{\text{\boldmath$\phi$}_{i}}( \lambda_T ) = \underset{ \text{\boldmath$\phi$} \in \mathbb{R}^{N+1} }{  \text{argmin} }  \ \left\{ \frac{1}{T} \sum_{t=1}^T \left(  y_{i,t+h} - \phi_{i,0} - \sum_{j=1}^N \text{\boldmath$\phi$}_{ij}^{\prime} \mathbf{X}_{j,t} \right)^2  + p_{\lambda_T}( | \text{\boldmath$\phi$}_{i}  | ) \right\}
\end{align}
where $p_{\lambda_T}(.)$ is a penalty function with a tuning parameter $\lambda_T > 0$ such that 
\begin{align}
p_{\lambda_T}( | \text{\boldmath$\phi$}_{i}  | ) &= \lambda_T  \sum_{j=1}^N | \phi_{ij} | \label{lasso} \\
p_{\lambda_T}( | \text{\boldmath$\phi$}_{i}  | ) &= \lambda_T   \sum_{j=1}^N \widehat{w}_j | \phi_{ij} | \label{adaptive}
\end{align}
where the weight is defined as $\widehat{w}_j = 1 / | \hat{\phi}^{ols} |^{\gamma}$ for some $\gamma > 0$. The penalty function given by expression \eqref{lasso} corresponds induces the standard lasso estimator while the penalty function given by expression \eqref{adaptive} corresponds to the adaptive lasso estimator. Both estimators are suitable for estimation and inference problems for high-dimensional models. In this paper we focus on the performance of these estimators when forecasting realized volatility measures which are considered to be highly correlated.

\newpage

The statistical literature provides various shrinkage functions which include; the standard Lasso penalty (see, \cite{tibshirani1996regression}), the Adaptive Lasso penalty (see, \cite{zou2006adaptive}), the SCAD penalty (see, \cite{fan2001variable}), the Elastic Net (see, \cite{zou2005regularization}) as well as the Priority Lasso (see, \cite{klau2018priority}). The choice of the penalty function allows the econometrician to consider the properties of the specification form which results to different sparse shrinkage estimators capturing this way different effects (such as, network, group or cross-section effects). Thus, the penalty function and the tuning parameter provides a feasible method for model and variable selection based on the nature of the data under examination.  

We consider that the time-series observations are indexed as $t =1,...,T$, for $N$ firms and $K$ predictors, where the number of predictors can be larger than the number of firms, such that $K \geq N$. Furthermore, since we operate within a high-dimensional modelling environment, the dimension of the vector of predictors,  $K$, is allowed to be larger than the number of time observations $T$. Assumption \ref{Assumption1} below provides regularity conditions that hold for the purpose of estimation and inference within our framework (see,  \cite{knight2000asymptotics}). 

\medskip

\begin{assumption}
\label{Assumption1}
Let $\epsilon_t$ be a sequence of innovations. The following moment conditions hold, which permit $X_{t}$ and $\epsilon_t$ to be weakly dependent, for all $i = \{ 1,...,N \}$. 
\begin{enumerate}
\item[A1.] $\{\epsilon_{i,t} \}_{t=1}^{T}$ is a homoskedastic \textit{martingale difference sequence} with $\mathbb{E} \left( \epsilon_{i,t} | \mathcal{F}_{t-1} \right) = 0$ and $\mathbb{E}\left( \epsilon_{i,t}^2  \right) = \sigma^2 \ \forall \ i = \{ 1,...,N \}$, where $\mathcal{F}_{t-1} = \left\{ X_t, X_{t-1},..., \epsilon_{t-1}, \epsilon_{t-2},... \right\}$.
\item[A2.] $\{X_t\}_{t=1}^{T}$ has at least a finite second moment, that is, $\underset{ t \in \mathbb{Z} }{ sup }\sum_{t=1}^{T} ||X_t||^{2 + s} < \infty$ where $s > 0$ and the following \textit{Weak Law of Large Numbers (WLLN)} holds
\begin{align}
\frac{1}{T} \sum_{t=1}^{T} X_t X_t^{\prime} \overset{p}{\to} \Omega_T \ \text{as} \ T \to \infty  \ \text{such that } \ \underset{ r \in  [0,1] }{ \text{sup} } \bigg\rvert  \frac{1}{T} \sum_{t=1}^{[Tr]} X_t X_t^{\prime} - r \Omega_T    \bigg\rvert = \mathcal{O}_p(1) 
\end{align}
where $\Omega_T$ a $K \times K$ non-stochastic finite and positive definite matrix.

\item[A3.] $\mathbb{E} \left( x_t \epsilon_t \right) = 0$ and $\mathbb{E} \left( x_{i,t} \epsilon_t  x_{j,t} \epsilon_t \right) = 0$, for all, $i,j,s,t \neq s$.
\end{enumerate}
\end{assumption}

\medskip

\begin{assumption}
\label{Assumption2}
Let $\mathcal{A}_T := \left\{ j : \hat{\beta}_j \neq 0 \right\}$ be the active set of predictors selected by \eqref{estimation}, where $\mathcal{A} = \left\{ 1,...,p_N \right\}$. The variable selection is consistent \textit{iff} 
\begin{align}
\underset{ T \to \infty }{ \text{lim} } \mathbb{P} \left( \mathcal{A}_T = \mathcal{A} \right) = 1.
\end{align}
\end{assumption}
Assumption  \ref{Assumption2} provides a condition for consistent variable selection. The shrinkage methodology is suitable for the high-dimensional setting. First, the properties of the Lasso norm provide a compromise between model fit and the curse of dimensionality. Second, the regularization technique ensures that the variance of the estimator is kept in low levels, since the particular methodology is built on the \textit{bet on sparsity} principle, which implies that the Lasso shrinkage estimator provides an one-step-go procedure that simultaneously estimates and selects the past RVs of the Cross-Section which are relevant to the forecasted values of RV of a certain firm from the pool of firms.

\newpage 

We assume that the tuning parameter is estimated via the cross-validation methodology. An optimal choice of $\lambda_T = \hat{\lambda}_T$ is essential to ensure that the Lasso estimator is consistent with an optimal rate of convergence (related proofs can be found in \cite{chetverikov2020cross}). Below, we present the cross-validation objective function. We follow the definition as presented by \cite{chetverikov2020cross}. Let $\kappa$ be a strictly positive integer, and let $\left( I_{k} \right)_{k=1}^{\kappa}$ be a partition of the set $\left\{ 1,...,T \right\}$, which implies that for each $k \in \left\{ 1,..., \kappa \right\}$. Moreover, for $k \neq s$, we consider mutually independent subsets, $I_k$ and $I_s$ such that $\bigcup_{k=1}^{\kappa} I_k = \left\{ 1,...,T \right\}$. Then, denote with $\Lambda_T$ the set of candidate values for $\lambda_T$. Therefore, to estimate the optimal value $\hat{\lambda}_T$ from the continuous compact set $\Lambda_T$, we define the  Leave-one-out lasso estimator for all $k=1,..., \kappa$ as below
\begin{align}
\label{leave.one.lasso}
\widehat{\text{\boldmath$\phi$}}^{(i)}_{-k}( \lambda_T ) = \underset{ \text{\boldmath$\phi$} \in \mathbb{R}^{N+1} }{  \text{argmin} }  \ \left\{ \frac{1}{T - T_{\kappa}} \sum_{ i \in I_{\kappa}} \left(  y_{i,t+h} - \phi_{i,0} - \sum_{j=1}^N \text{\boldmath$\phi$}_{ij}^{\prime} \mathbf{X}_{j,t} \right)^2  + p_{\lambda_T}( | \text{\boldmath$\phi$}_{i}  | ) \right\}
\end{align}
Specifically, the Leave-one-out lasso estimator, given by expression \eqref{leave.one.lasso}, is the Lasso estimator which corresponds to all observations excluding those in $I_k$, where $n_T = | I_k |$ is defined to be the size of the subsample $I_k$. The Leave-one-out lasso estimator has the same properties as the original Lasso estimator, which means that the optimization problem has a unique solution which converges to the true value of the estimator with probability one. Then, the cross-validation choice of  $\lambda_T$ is given by
\begin{align}
\hat{\lambda}_T^{(i)} = \underset{ \lambda_i \in \Lambda_T
 }{ \text{argmin} } \ \ \sum_{k=1}^{\kappa} \sum_{ t \in I_{\kappa}} \left(  y_{i,t+h} - \sum_{j=1}^N \widehat{\text{\boldmath$\phi$}}^{(i)}_{-k}( \lambda_T ) \mathbf{X}_{j,t} \right)^2
\end{align}   
This procedure is repeated for each of the $N$ cross-section specifications we have, therefore we obtain a sequence of cross-validation parameters $\left\{ \hat{\lambda}_T^{(i)} \right\}_{i=1}^N$. Related asymptotic results in the literature show that the Lasso estimator has good accuracy performance  for both estimation and forecasting purposes, for large values of $p_N$, assuming that the true population value $\beta$ is sparse. In particular, under Assumption \ref{Assumption1}, one can prove the asymptotic normality of the lasso estimator (see, \cite{knight2000asymptotics}). 

In practise, the forecast evaluation study implies a time-varying LASSO scheme which is an essential setting for obtaining out-of-sample forecast error sequences, using a suitable loss function (e.g., quadratic, absolute value etc). Within our framework we assume that the choice of the loss function does not affect the asymptotic theory in consistently estimating the sparse lasso estimator for the time-varying case. For instance, \cite{kapetanios2018time} introduce a Lasso-type estimator for high-dimensional linear models with time-varying parameters, which is similar to our proposed framework. Moreover, Theorem 1 in the particular paper proves that the time-varying lasso is consistently estimated for a given cross-validation parameter $\lambda_T$. 

The use of the simple Lasso selection methodology for the purposes of forecasting RV has certain limitations. The particular shrinkage method  might omit active variables since is based on the strength of the individual variables as opposed to the strength of the groups of input variables. In order to account for this effect, we consider additional penalty functions as robustness checks such as the Adaptive Lasso penalization methodology as explained above.

\newpage

\subsection{Realized Volatility Model Specifications}

\paragraph{Preliminary Theory} In this section, we present the related background to the measure of Quadratic Variation. Let the stock price denoted with $X_t$, then it can be proved that the logarithmic price process follows an It$\hat{\text{o}}$ semi-martingale given by
\begin{align}
\label{Q.Veq} 
X_t = X_0 + \int_0^t \mu(s) ds + \int_0^t \sigma(s) dW(s) + J(t) 
\end{align}
where $\mu(s)$ is a predictable process and $\sigma(s)$ is the instantaneous volatility assumed to be stationary, latent and stochastically independent of $W(s)$, the standard Brownian motion and $J(t)$ is a finite-activity pure jump process, such as, $J_t = \sum_{j=1}^{L_t} \kappa_j$ where $L$ is a counting process and the $\kappa_j$'s are random variables for the size of the jump of the stochastic process\footnote{Notice that the unobservable nature of the QV of asset prices requires to impose strong assumptions for the robust measurement of asset volatility. Specifically, assuming that the empirical process that describes the generation of QV  involves no measurement error, microstructure noise as well as any jump components, then the time series sequence of the observable RV is a sufficient proxy of the latent volatility. Thus, to account for the above effects when doing statistical inference with the econometric environment described in Section \ref{Section2.1}, we need to impose specific regulatory conditions.}. We then define the quadratic variation, (QV), of the price process to be $\displaystyle QV(t) = \int_0^t \sigma^2(s) ds + \sum_{0 \leq s \leq t} \big[ \Delta J(s) \big]^2$. Then, the volatility proxy is estimated as below
\begin{align}
\displaystyle RV_t^{(d)} = \sum_{j=0}^{M-1} r^2_{t-j.\Delta} \  \text{and} \ RV_{t,M} \overset{p}{\to} IV_t + JV_t \ \text{as} \ M \to \infty,
\end{align}
where $\Delta = 1d / M$ the sampling frequency of the intra-day returns. 
\begin{assumption}
\label{Assumption3}
The process $\left( X_i \right)_{ 1 \leq i \leq M }$ that satisfy \eqref{Q.Veq} is adapted and locally bounded and the jump components are given by 
\begin{align}
\sum_{0 \leq s \leq t} \Delta X_{i,s} = \int_0^t \int_{\mathbb{R}}  \delta_i \left( \omega, s, z  \right) \mu \left( \omega, ds, dz \right).
\end{align} 
\end{assumption}

\subsubsection{Realized Volatility Specifications}  
  
The econometric framework proposed in Section \ref{Section2.1}, is applied to a Large Cross-section of RV measures by replacing the dependent variable in the nonlinear cross-section model specifications with $y_{i,t+1} \equiv \text{RV}_{i,t+1}$ and the set of predictors with $\mathbf{X}_{j,t}= \left[ \text{RV}_{j,t} \ \vdots \ \overline{\text{RV}}^{w}_{j,t} \ \vdots \ \overline{\text{RV}}^{m}_{j,t} \right]$ using the following temporal volatility filters
\begin{small}
\begin{align}
\label{volatility.filters}
\text{RV}_{i,t+h}= \sum_{l=1}^h \frac{1}{h} \text{RV}_{i,t+l} , \ \overline{\text{RV}}_{j,t}^{w}= \frac{1}{5} \sum_{l=0}^4 \text{RV}_{j,t-l} \ \text{and} \ \ \overline{\text{RV}}_{j,t}^{m}= \frac{1}{22} \sum_{l=0}^{21} \text{RV}_{j,t-l} 
\end{align}
\end{small}
The benchmark model is estimated using the standard least squares estimation method which does not affect estimation accuracy due to the low-dimension of the estimation problem. On the other hand, a consistent estimator for mixed frequency data can be obtained under further regularity conditions (see, \cite{andreou2010regression}). Furthermore, we examine alternative specifications to the HAR model based on the Lasso shrinkage norms we consider in this paper.

\newpage

\begin{table}[h!]
  \centering
  \caption{Realized Volatility-based Model Specifications}
    \begin{tabular}{|c|l|}
    \hline
    \textbf{Model } & \multicolumn{1}{c|}{\textbf{Model Specification }} \\
    \hline
    \multirow{3}[0]{*}{\textbf{HARQ}} & \multicolumn{1}{l|}{\multirow{3}[0]{*}{$y_t = \beta_0 + \left( \beta_1 + \beta_{1Q} x^{1/2}_{t-1} \right) y_{t-1} + \beta_1 y_{t-1|t-5} + \beta_3 y_{t-1|t-22} + u_t$}} \\
          &  \\
          &  \\
    \hline
    \multirow{3}[0]{*}{\textbf{HAR-J}} & \multicolumn{1}{l|}{\multirow{3}[0]{*}{$y_t = \beta_0 + \beta_1 y _{t-1} + \beta_2 y_{t-1|t-5} + \beta_3 y_{t-1|t-22} + \beta_j J_{t-1} + u_t$}} \\
          &  \\
          &  \\
    \hline      
    \multirow{3}[0]{*}{\textbf{CHAR}} & \multicolumn{1}{l|}{\multirow{3}[0]{*}{$y_t = \beta_0 + \beta_1 x _{t-1} + \beta_2 x_{t-1|t-5} + \beta_3 x_{t-1|t-22} + u_t$}} \\
          &  \\
          &  \\
    \hline
    \hline
   \end{tabular}%
  \label{tab:addlabel}%
\end{table}%

\begin{itemize}
\item \textbf{HARQ model}: is a specification which incorporates a measurement error correction to the parameters, where $y_t \equiv \text{RV}_t$, $x_t \equiv \text{RQ}_t$ and RQ$_{t-1|t-k} \equiv \frac{1}{k} \sum_{j=1}^k \text{RQ}_{t-j}$.   

\item \textbf{HARQ model}: is a specification which accommodates a measurement error correction. The model provides a robust identification of the jump component which is largely unpredictable. The variables represent, $y_t \equiv \text{RV}_t$, $x_t \equiv \text{RQ}_t$ and J$_{t} \equiv \text{max}[ \text{RV}_t - \text{BPV}_{t} ,0]$ and BPV is defined by BPV$_t \displaystyle \equiv \mu_1^2 \sum_{i=1}^{M-1} | r_{t,i} | | r_{t,i+1} |$ with $\mu_1 = \sqrt{ 2 / \pi} = E[ | Z| ]$ and $Z \sim N(0,1)$. 

\item \textbf{CHAR model}: is a specification which relies on the decomposition of the total variation of the unobserved Quadratic Variation into a continuous part approximated via the Bi-power Variation (BPV) measure of  \cite{barndorff2004power} which provides a consistent estimate of the continuous variation in the presence of jumps. 

\end{itemize}

The HARQ model is an improved model specification of the standard HAR model which incorporates a measurement error correction to the parameters. Specifically, realized volatility is a proxy variable of the latent quadratic variation, which is thus unpredictable and can affect the forecasting ability of RV measures. For the HARQ model the measurement error correction\footnote{Note that, the error correction mechanism works using the following intuition: The econometric specification has the form $y_t = \beta_t x_{t-1} + u_t$, and we further assume that the measurement error affects the evolution of the time-varying model coefficient in a linear manner such that, $\beta_t = \delta_0 + \delta_1 \sqrt{ z_{t-1}}$, which implies that $y_t = \delta_0 x_{t-1} + \delta_1 w_{t-1} + u_t$, with $w_t = \sqrt{z_{t-1}} x_{t-1}$.} is applied only to the one day lagged RV, however correcting the other two components as well implies the HARQ-F specification. Both model specifications, are considered as simple examples of how the inclusion of regressors with different degree of persistence has a direct impact on forecasting. In particular, \cite{hansen2014estimating} propose an instrumental variable methodology which estimates the degree of persistence in time series measured with error. For the purpose of this paper we focus on the forecast evaluation of the alternative HAR specifications (such as HARQ, HAR-J and CHAR). On the other hand, the HARJ model includes a measure of the jump variation as an additional explanatory variable in the standard HAR model, following the proposition of \cite{andersen1998answering}, while the CHAR model includes only measures of the continuous variation.

\newpage

\begin{remark}
Note that the volatility filters introduced by expression \eqref{volatility.filters} are constructed using temporal aggregation. In this paper, we consider the conventional approach which employs the model specification subject to the temporal aggregation restriction of an equal weighting scheme. For instance, \cite{andreou2010regression} consider a flexible data-driven aggregation scheme and a suitable testing methodology is proposed to examine the effect of the weighting scheme on the relative efficiency of the estimators. 
\end{remark}    
In summary, the additional specifications such as the HARQ or the HAR-J models, provide relative improvements in terms of forecast ability that depends on both the nature of the measurement error due to increased levels of market volatility. For example, according to \cite{liu2015does}, if the measurement error is constant over time, the HARQ model reduces to the standard HAR model. When the measurement error is highly heteroscedastic, the HARQ model provides the greatest forecast accuracy improvements in comparison to other benchmark specifications.  These models incorporate features of the RV such as jump volatility, measurement error correction and the continuous variation component. Moreover, the model specifications estimated via the shrinkage norms can provide statistical evidence of the sensitivity of the forecast evaluation testing methods to the model specification and specifically to the proxy risk measures as well as how these features affect the cross-sectional dependence.


\section{Predictive Accuracy Testing for Cross-Sections}
\label{Section3}

We motivate the econometric framework for the forecast evaluation of RV which allows to assess whether using the cross-section significantly improves the predictive accuracy of RV measures for particular firms. The main challenge for the forecast evaluation exercise of risk measures such as the stock volatility remains the fact that even though are well-defined in models, are not observable even ex-post. Specifically, \cite{li2018asymptotic} propose related asymptotic theory for the application of forecast evaluation tests to measures of RV. The proposed framework, is suitable for testing predictive accuracy of a target variable which is unobservable such as in the case of volatility. The authors propose to use a proxy of the latent variable using high-frequency (intra-day) data. Moreover, the papers of \cite{patton2011data}, \cite{liu2015does} and \cite{bollerslev2016exploiting} provide extensive examination of various realized volatility estimators under different modelling conditions. A robust framework which tackles the main econometric aspects for forecasting under model uncertainty are presented by \cite{hirano2017forecasting}.

The theoretical background to the predictive accuracy is driven by the framework of decision theory\footnote{Note that estimation under quadratic loss is proposed with the seminal paper of James and Stein 1961}. The decision maker or forecaster has some prior information regarding the parameter of interest $\theta \in \Theta$ but has only observed a trajectory of these data. In particular, the forecaster has access to competing point forecasts, therefore a common practise for evaluating their performance is via the use of a loss or score function. In practise, the incentive of the decision maker is the minimization of the expected loss:
\begin{align}
\mathcal{R} \left( \hat{Y}_{t+1}  \right) = \mathbb{E} \left[ \mathbb{E} \left[ L \left( \hat{Y}_{T_0+1}, Y_{T_0+1} \right)  \right] \big| \theta = \Theta  \right]
\end{align}    
where $L_{i,t+1} \equiv L \left( \hat{Y}_{T_0+1}, Y_{T_0+1} \right)$ denotes the loss function.    

Within our framework, the out-of-sample forecasts are generated by the cross-section predictive regression specification given by expression \eqref{specification}. The forecasting scheme can be constructed using a rolling or expanding window. For instance, when considering a recursive window parameter estimation, the forecast sequences are obtained  by regressing $y_{s+h}$ on $\mathbf{X}_{s}$, for $s = 1,...,t$, where $t \in \{ 1,...,T_0 \}$. Then, we obtain the Lasso estimates $\widehat{\text{\boldmath$\phi$}}_{it}$ as expressed by \eqref{estimation} and using $\hat{y}_{t+h|t} ( \widehat{\text{\boldmath$\phi$}}_{it} ) = \widehat{\text{\boldmath$\phi$}}^{\prime}_{it} \mathbf{X}_{t}$, $h-$period ahead forecasts are constructed (e.g., with $h \equiv 1$). To apply the tests of Equal Predictive Accuracy, forecasts from a constrained model are compared with forecasts from an unconstrained model. Let $\mathcal{M}^{(i)}_0$ be the benchmark model (standard HAR model) for each firm separately and let $\mathcal{M}^{(i)}_1$ be the set of all possible nonlinear cross-section regressions from the family of model specifications we consider in Section \ref{Section2.1} and Section \ref{Section2.2}. The set of model specifications in $\mathcal{M}^{(i)}$ can be nested, nonnested, and overlapping.

\subsection{Tests of Equal Predictive Accuracy}
\label{Section3.2}

We denote with $\hat{y}_{t+h|t}$ the $h-$period ahead forecast based on the benchmark model and $\tilde{y}_{t+h|t}$ the $h-$period ahead forecast based on the nonlinear cross-section model which belongs to the family of models $\mathcal{M}_1$. Then, under the mean squared error (MSE) loss function we estimate the forecast sequences for the various models and construct the equal predictive accuracy (EPA) tests, under the null hypothesis of equal predictive power. A test of equal predictive accuracy takes the following form: 
\begin{align}
\mathbb{H}_0: \mathbb{E} \big[ y_{t+h} - \hat{y}_{t+h|t} \big]^2 =  \mathbb{E} \big[ y_t - \tilde{y}_{t+h|t} \big]^2
\end{align}

The literature of predictive accuracy testing has been introduced and advanced via the proposed framework of \cite{diebold2002comparing}, \cite{west1996asymptotic} and \cite{clark2007approximately}. Extensions of the particular tests have been examined in terms of asymptotic and finite sample results which depend on the testing conditions and assumptions of the econometric framework one considers. Furthermore, one can also consider the forecasting procedure of superior predictive accuracy, (SPA), proposed by \cite{white2000reality} for comparing multiple forecasting models. Last but not least, while these tests are considered to be unconditional EPA testing methodologies, the conditional EPA test proposed by \cite{giacomini2006tests} focuses on evaluating the accuracy of the forecasting method rather than the accuracy of the forecasting model in comparison to other tests. 

In this paper, we focus on the use of the unconditional and conditional EPA tests of predictive accuracy for the out-of-sample fitness of the two models under consideration. The EPA testing methodology requires the comparison of a benchmark model, under the null hypothesis and the forecast model, under the alternative hypothesis, which includes additional predictors. Specifically, considering a simple forecasting accuracy test for an in-sample evaluation, then we have $\{ \hat{y}_{it}  \}_{t=1}^T$ and $\{ \hat{y}_{jt}  \}_{t=1}^T$ the two forecasts of the time series $\{ y_{t}  \}_{t=1}^T$ with associated forecast errors be $\{ \epsilon_{it} \}_{t=1}^T$ and $\{ \epsilon_{jt} \}_{t=1}^T$. Then, using a loss function with certain functional form (e.g., quadratic loss) of the form $f( y, \hat{y}_{it} ) \equiv f( \epsilon_{it} )$, the testing hypothesis under examination is 
\begin{align}
\mathbb{H}_0:       \mathbb{E}[ f( \epsilon_{it} ) ] = \mathbb{E}[ f( \epsilon_{jt} ) ] \ \ \text{versus} \ \ \ \mathbb{H}_1: \mathbb{E}[ f( \epsilon_{it} ) ] > \mathbb{E}[ f( \epsilon_{jt} ) ]
\end{align}
which implies $\mathbb{E}[ d_t ] = 0$ with $d_t = f( \epsilon_{it} ) - f( \epsilon_{jt} )$. Therefore, a rejection of the null hypothesis H$_0$ implies that we accept that the benchmark model has higher MSE.

\newpage

Accepting the alternative hypothesis is interpreted as a statistical significant preference of the decision maker towards the extended model since it has smaller MSE which is an indication that the model under the alternative hypothesis has better predictive accuracy for the Realized Volatility of the Cross-Section. Thus, the proposed forecast evaluation of the RV measures is a model-based forecasting procedure, that is, we directly model the moment $\mathbb{E}[ y_{ t + \tau} | \mathcal{F}_t ]$. 

Related background theory to testing methodologies for forecast evaluation, can be found in \cite{west2006forecast}. The particular paper covers both nested and non-nested models and presents the corresponding asymptotic theory, for out-of-sample forecast evaluation environments. Suitable model performance criteria include for example the Mean Square Prediction Error (MSPE). Moreover, the structure of the competing models has a direct impact on the asymptotic distribution and critical values of the test statistic. For instance, \cite{mccracken2007asymptotics}, considers OOS predictive ability tests of nested parametric models. \cite{clark2014tests} presents the adjusted EPA testing methodology and asymptotic results for overlapping models. Furthermore, \cite{busetti2013comparing} provide an examination of the various tests in the literature for comparing forecast accuracy via a Monte Carlo study. Below, we present in the form of examples the main tests we consider in this paper. 

\begin{example}(Diebold $\&$ Mariano Test)
Assume we obtain the forecast error sequences within the out-of-sample period, i.e., $t=R,..,T$, then 
\begin{align}
 y_t = X'_{1t} \beta_1^{*} + \epsilon_{1t} \ \text{and} \ y_t = X'_{2t} \beta_2^{*} + \epsilon_{2t}
\end{align}
We define the one-step ahead (i.e.,$\tau=1$) forecast errors and sample MSPEs as below
\begin{align}
\hat{\epsilon}_{1,t+1} &= ( y_{t+1} - X'_{1,t+1} \hat{\beta}_{1,t} ), \ \ \ \hat{\epsilon}_{2,t+1} = ( y_{t+1} - X'_{2,t+1} \hat{\beta}_{2,t} ) 
\end{align}
where
\begin{align}
\hat{e}_1^2 &= \frac{1}{P} \sum_{t = R}^T \hat{\epsilon}_{1,t+1} , \ \ \ \ \ \ \ \ \ \ \ \ \hat{e}_2^2 = \frac{1}{P} \sum_{t = R}^T \hat{\epsilon}_{2,t+1}
\end{align}
Under the MSPE performance measure the object of interest for EPA testing is the difference between sample MSPEs $\hat{e}_1^2$ and $\hat{e}_2^2$.
We define the following quantities, 
\begin{align}
 \hat{f}_t \equiv \hat{e}_{1,t} - \hat{e}_{2,t} \ \text{and} \ \bar{f} \equiv \frac{1}{P} \sum_{t=R}^T \hat{f}_{t+1} \equiv   \hat{e}^2_{1,t} - \hat{e}^2_{2,t}
\end{align}
\end{example}
\begin{example}(Clark $\&$ West OOS-F test)
\begin{align}
\hat{e}^2_1 &= \frac{1}{P} \sum_{t=R}^T ( y_{t + \tau} - \hat{y}_{1, t + \tau})^2 \\
\hat{e}^2_2 &= \frac{1}{P} \sum_{t=R}^T ( y_{t + \tau} - \hat{y}_{2, t + \tau})^2 
\end{align}
Under, the null hypothesis, we have that MSPE$_1$ = MSPE$_2$. The particular test statistic adjusts for the possible inefficient estimation of Model 2 due to sample bias of population parameters by adjusting for the difference between the MSPE of the two models. Thus, the test statistic of CW is computed using the MSPE-adjusted, denoted as $\hat{f}_{t+1} = \hat{e}^2_{1,t+1} - [ \hat{e}^2_{2,t+1} - ( \hat{e}_{1,t+1} -  \hat{e}_{2,t+1})^2 ]$ with the corresponding sample average, denoted as, $\bar{f} = \frac{1}{P} \sum_{t=R}^T \hat{f}_{t+1}$. Thus, the CW test statistic for the case $\tau = 1$ (one-period ahead) is defined as below:
\begin{align}
 \sqrt{P} \frac{ \bar{f} }{ \sqrt{ \text{Var}[ \hat{f}_{t+1} - \bar{f} ] } }
\end{align}
\end{example}

\begin{example}(Giacomini $\&$ White test)
Under the null hypothesis the conditional EPA  test of Giacomini and White requires that 
\begin{align}
\mathbb{H}_0: \mathbb{E} \left[ L_{t+ \tau}( Y_{t+ \tau}  ,  f_t( \hat{\beta_1}) ) -  L_{t+ \tau}( Y_{t+ \tau}  ,  g_t( \hat{\beta_2}) ) \vert \mathcal{G}_t   \right] \equiv \mathbb{E} \left[  \Delta L_{m,t+\tau}  \vert  \mathcal{G}_t  \right] = 0 \ \text{a.s} 
\end{align}
where $\tau$ the period ahead to be forecasted, $m$ the rolling window size and $L_{t+ \tau}(.)$ the loss function. The Wald-type statistic for conditional EPA testing (one-step ahead) of the competing models is given by the following expression:
\begin{align}  
T_{m,n}^h  = n \left( \frac{1}{n} \sum_{t=m}^{T-1} h_t \Delta L_{m,t+1}  \right)^{'} \hat{\Omega}^{-1}_n \left( \frac{1}{n} \sum_{t=m}^{T-1} h_t \Delta L_{m,t+1}  \right) 
           \equiv n \bar{Z}^{'}_{m,n} \hat{\Omega}^{-1}_n \bar{Z}^{'}_{m,n}
\end{align}
where $\bar{Z}^{'}_{m,n} = \frac{1}{n} \sum_{t=m}^{T-1} Z_{m, t+1}$, $Z_{m, t+1} =  h_t \Delta L_{m,t+1}$, and   $\hat{\Omega}_n \equiv \frac{1}{n} \sum_{t=m}^{T-1} Z_{m, t+1} \times Z^{'}_{m, t+1}$, is a $q \times q$ matrix that consistently estimates the variance of $Z_{m, t+1}$. 
For example, if we consider that the test function $h_t$ is given by $h_t = (1, \Delta L_{m,t} )^{'}$ then we have that the regressor matrix is given by
\begin{align}
Z_{m,t+1} = (1, \Delta L_{m,t} )^{'} \Delta L_{m,t+1} = \left( \Delta L_{m,t} \ , \ \Delta L_{m,t} \Delta L_{m,t+1} \right). 
\end{align}  
Thus, the proposed test with statistical significance $\alpha$ is constructed by rejecting the null hypothesis of equal conditional predictive ability whenever $T_{m,n}^h > \chi^2_{q, 1 - \alpha}$, where $\chi^2_{q, 1 - \alpha}$ is the $(1 - \alpha)$ quantile of a $\chi^2_{q}$ distribution and $q$ the number of instruments used in the $h_t$ test function. For the asymptotic justification of the test under the null hypothesis, see Theorem 1 in \cite{giacomini2006tests}. Note also the GW test is suitable for a rolling window forecasting scheme and not via expanding window.  Furthermore, the authors consider the implications of rejecting equal conditional predictive ability and describe a method for adaptively selecting at time $T$ a forecasting method for $T+\tau$, which is forecast selection mechanism. The basic idea of this methodology is that rejection occurs because the test functions $\{ h_t \}$ can predict the loss differences $\{ \Delta L_{m,t+\tau} \}$ out-of-sample, which suggests using $h_T$ to predict which method will yield lower loss at $T+\tau$. The methodology involves the following steps
\begin{itemize}
\item Regress $\Delta L_{m,t+1} = L_{t+\tau}( Y_{t+\tau}, \hat{f}_{t, m_f} ) - L_{t+\tau}( Y_{t+\tau}, \hat{g}_{t, m_g} )$ on $h_t$ .

\item The approximate $\delta^{'} h_T \approx E[ \Delta L_{m,t+1} | \mathcal{F}_t ]$ motivates the decision rule: use $g$ if $\delta^{'} h_T > c$ and use $f$ if $\delta^{'} h_T < c$, with $c$ a user-specified threshold e.g., $c=0$.   
\end{itemize} 
\end{example}

\begin{remark}
The use of the above three forecast evaluation tests can be useful in examining different econometric conditions depending on the assumptions of the time series under consideration. However, we might expect that some of the tests e.g., the \cite{clark2007approximately} and the \cite{giacomini2006tests} tests provide similar results regarding the forecast evaluation of the Cross-Sectional RV across the forecast models of all firms. 
\end{remark}

\newpage

Within our framework, the variable of interest is the Realized Volatility of a firm in the Cross-Section. Better forecasts of the Cross-Sectional RV can provide significant improvements in estimating risk related to the cross-section which as a result helps to provide better estimates for the expected returns of the Cross-Section. In Section \ref{Section3.3}, we examine in more details some specific features which are related to the Cross-Section.

\subsection{Cross-Section Dynamics}
\label{Section3.3}

In this Section we aim to identify more rigorously the Cross-Section dynamics by examining in more details the nature of the data, the variable selection methodology as well as possible modifications in the tests of predictive accuracy which can provide a power enhancement mechanism. More specifically, the nature of the data we consider are highly correlated, and this is the case for both the firm-specific predictors (benchmark model) as well as the cross-section specific predictors (forecast model). Furthermore, under the Assumption \ref{Assumption1} - \ref{Assumption3} and the estimation procedure described   in Section \ref{Section2.1} and \ref{Section2.2}, we can accommodate the Elastic net shrinkage which is suitable for correlated predictors.  

In particular, the elastic net mixes both the $\mathcal{L}_1$ and $\mathcal{L}_2$ norms for the penalty term, 
\begin{align}
\widehat{\text{\boldmath$\theta$}_i}( \lambda_T ) = \underset{ \text{\boldmath$\theta$} \in \mathbb{R}^{N+1} }{  \text{argmin} }  \ \left\{ \frac{1}{ T } \sum_{ t = 1 }^T \left( y_{i,t+h} - \theta_{i,0} - \sum_{j=1}^N \text{\boldmath$\theta$}_{ij}^{\prime} \mathbf{X}_{j,t} \right)^2  + p_{\lambda_T}( | \text{\boldmath$\theta$}_{i} | ) \right\}
\end{align}
where $\text{\boldmath$\theta$}_{i}$ is the corresponding elastic-net estimator and the penalty function is given by
\begin{align}
p_{\lambda_T}( | \text{\boldmath$\theta$}_{i} | ) = \lambda  \big[ \eta \norm{ \text{\boldmath$\theta$}_{i} }_1 + (1 - \eta) \norm{ \text{\boldmath$\theta$}_{i} }^2_2  \big],  \ \ \ \text{where} \ \ \eta = 0.5.
\end{align}
Moreover, as we observe also via the Empirical application of the paper, it seems that when we choose the elastic net shrinkage estimator which accommodates the highly correlated nature of the data, then the tests of equal predictive accuracy produce higher rejection rates of the null hypothesis of equal predictive accuracy. In other words, this implies that when we include the feature of high correlation among the predictors then, it is more likely to find statistical evidence of predictive ability of the cross-section as reflected from the comparison of the forecasts produced via the benchmark and the forecast model. In particular, the elastic-net achieves minimax optimality (see, \cite{zou2005regularization}) by standardizing each predictor with the pairwise correlations, providing this way a stabilized version of the standard Lasso estimator.       

Since we operate within a high dimensional framework, the use of the shrinkage estimators implies that the loss function used to estimate the model parameters are different that the loss function used to measure predictive ability (assuming a quadratic loss function), which can introduce a dependence of the limiting distribution to the data generating process. However, we can observe that combining additional information and parameter constraints with Lasso (as it is the case of the Elastic net estimator), can further improve the predictability of the firm-specific realized volatility using the Cross-Sectional information set (see, \cite{fan2015power}).   

\newpage

\section{Empirical Application}

The empirical application aims to provide supporting evidence of the theoretical background. In particular, the Augmented HAR model via the cross-section which is estimated via variable selection norm allows to construct forecasts based on the predictors incorporated in the cross-sectional information set. Moreover, the shrinkage estimators we consider in the paper provide a suitable framework in which we can examine how the degree of sparsity in variable selection affects the cross-section based forecasting methodology across firms. In summary, in terms of the econometric importance of this setting, is that our empirical application allows to investigate how the effect of predictors with mixed integration order affect the predictive accuracy of the forecast model which is based on the cross-sectional augmentation process.

\subsection{Forecasting Realized Volatility in Large Cross-Sections} 

In the first empirical application of this paper, we estimate the above HAR specifications and compare their performance in terms of statistical tests of equal predictive accuracy which is the proposed testing methodology we implement in this paper. The literature has indeed documented that the HAR-J and CHAR models perform slightly better than the standard HAR model (see, e.g, \cite{bollerslev2016exploiting}). Furthermore, additional HAR specification can be constructed such as the Semivariance-HAR (SHAR) model which decomposes the total variation due to the negative and positive intraday returns respectively (see, \cite{barndorff2008measuring}). Moreover, \cite{barndorff2008designing} propose a noise reduction methodology for RV measures using kernel functions. Our framework considers the cross-sectional dependence in second-order moments of high frequency data.
For instance, the goal of a forecaster  is to investigate the presence of predictability in the RV measures of a certain firm based on the cross-sectional jump components. We argue that the phenomenon of episodic predictability might be explained by periods in which the cross-sectional jump components are consistently selected via the shrinkage estimation procedure in comparison to other periods of reduced market uncertainty (see, \cite{jurado2015measuring} and \cite{baker2016measuring}). This approach of considering discontinuities in the dynamics of aggregate market risk measures is examined by \cite{li2017jump} who construct a formal econometric framework for jump regressions (see, also \cite{li2017Robust}).      
 
\subsubsection{Data Description} 

We consider high frequency data such as intra-day returns for a cross-section of firms. In particular, we use a large Cross-Section of RV, that is, the individual stock dataset of \cite{bollerslev2016exploiting}, which allow us to explore the predictive ability of the Cross-Section in forecasting RV. More specifically, the individual stock risk measures of RV for this dataset are based on the TAQ database. The sample starts on April 21,1997 and ends on December 31, 2013 accounting for a total of 4202 daily observations. For validation purposes we replicate some of the results already presented on BPQ, as we explain below, however the focus of our paper is the Cross-Sectionally Augmented HAR model and its variations which is not considered by BPQ. Table \ref{Table1} provides a standard set of summary statistics for the daily RV. 

\subsubsection{In-sample performance evaluation}

We begin our analysis by considering the full-in sample results. Table \ref{Table1} provides a summary of the main in-sample estimation results. We consider the average measures of $R^2$ and MSE across individual stocks for both the benchmark and forecast models.  In addition to the HAR model, we consider the HARQ and HARQ-F models which accommodates a correction for the measurement error in the estimation of the high-frequency realized volatility. In addition to these two model specifications we consider the HAR-J model which allows to examine the effect of Cross-Sectional jump components in the RV forecasting as well as the CHAR model which reflects a continuous time representation of QV. 

\subsubsection{Out-of-sample performance evaluation}

We consider two forecasting sample sizes (rolling window versus expanding windows with window size of 252 and 1000. Given an estimation sample of size $T$, we focus on forecasting $y_{T+1}$ using the proposed forecast methods. We then evaluate each forecast strategy in terms of the one-step-ahead forecast errors associated with $\hat{y}_{T+1}$. The out-of-sample estimation study aims to present the results of the forecast evaluation of the RV via the Cross-section. As a matter of fact, all estimated HAR models in this setting are considered to be time-varying coefficient models since the estimation procedure is repeated in each estimation window. Note that,  \cite{inoue2017rolling} proposes a methodology for optimal choice of the rolling window, which we can accommodate.  

We use the standard notation and terminology of forecast evaluation (e.g., see \cite{west2006forecast}) to describe the econometric framework we operate. Firstly, we split the full sample into two mutually exclusive subsets. The first $R$ observations are used as the estimation period and the next $P$ observations are used as the forecast evaluation period (i.e., $T+1 = R + P$), where $T = 4202$. In particular for the rolling window forecasting scheme we consider the effect of window size on the performance of the EPA tests within our framework. Thus, we consider for example a rolling window of size e.g., $P=1000$ which produces $3202$ forecast error sequences. To do this, in each rolling window we exclude the last observation we fit the benchmark and forecast models based on the remaining observations and then we forecast the 1000 observation. This allows to obtain a forecast error sequence which can be used to construct the EPA tests. Moreover, we can also consider the Expanding Window (recursive estimation) forecasting scheme in which the first observation of the window always remains to be the first observation of the full sample while the estimation window expands from the initial size of the window until the $T-1$ observation. 

To be able to have comparable results across the different EPA tests we use the same window size for the construction of the benchmark and forecast models (e.g., for the rolling window forecasting scheme). Attention is given to cases where the tests incorrectly accept the a false null hypothesis which can interpreted as a shortcoming of the test to indicate statistical evidence of non-equal predictive accuracy among the benchmark model and forecast model when actually one exists. The empirical application of the paper relies on daily realized volatilities estimated from intra-day data at 5-minute sampling frequencies (e.g., see \cite{liu2015does}). In particular when we use RV measures across different financial markets an important issue of concern when forecasting RV at time $t$, is to ensure that there is no contamination with  $t+1$ information about asset $i$ across the Cross-Section. For example, if one financial market is lag by one day then forecasting the RV for any financial market in the Cross-Section could lead to such a scenario as in \cite{bollerslev2018risk}. This particular issue is related to the synchronization/alignment issue of intra-daily stock returns as discussed in the related literature. Moreover, in this paper we also extend the above Lasso shrinkage specifications to the case of nested environments. In order to preserve the nested environment in related forecast evaluation applications we use the Priority Lasso (see, e.g., \cite{klau2018priority}) as a solution to the problem of selecting covariates from the Cross-Section in addition to the three own covariates of the firm under examination.

\subsubsection{Main Empirical Results}

The empirical application of the paper provides statistical and economical evidence of forecasting improvements in RV using the cross-sectional information set. In particular, the effect of excess market volatility fluctuations is directly reflected via the use of Large Cross-Sections which incorporate both over valued and under valued stocks. However, this is a phenomenon which clearly depends on the nature of the forecasting scheme, the model specification as well as the testing methodology. Nevertheless, we can draw some distinctive conclusions via our empirical application. Comparing the OOS results estimated via a window size of 252 time observations (see, Table \ref{Table2}) versus a window size of 1000 time observation (not reported here) it seems that model and parameter uncertainty in the first case has less negative impact on the forecast evaluation tests in terms of non-rejections of the null hypothesis. For example, in those cases for which the null hypothesis is non-rejected is worth looking at the characteristics of the specific firm, which appear to be firms with high value supplied chains and this also supports the claim that such firms which are highly interconnected in terms of the quality of their economic transactions and not necessarily in terms of their ties with the financial markets seem to indicate that their future stock volatility is not directly affected via the cross-section at least as seen by the forecast evaluation exercise of this paper. For an estimation window with size length 1000, the forecast evaluation tests may give a different picture due to the high correlation of asset prices to business cycle events included in the specific window. 

Incorporating the cross-section for forecasting RV has intuitive meaning also beyond the methodological framework of this paper. In particular, forecasting RV using the cross-sectional information set aims to shed light on economic theory related aspects, such as the construction of an expectations-augmented curve for volatility as well as to provide statistical evidence for an expectations driven volatility. Our framework of valuing stock risk levels via the measure of realized volatility is in accordance to the fundamental theorem of asset pricing which ensures that a market admits no arbitrage provided that we evaluate the aggregate risk via a martingale measure. Our proposed methodology exactly aims to provide a robust forecasting measure of realized volatility and is done by the elimination of individual preferences using the cross-sectional distribution of investment preferences reflected to the measure of Quadratic variation and approximated by the measure of RV.  

Moreover, the RV predictability as captured by the Cross-Sectional shrinkage norms closely resembles features of dynamic factor models with the advantage of not having to worry about the consistency of such estimators. In other words, the variable selection via the shrinkage norms captures a small number of unobserved factors aiming to explain the co-movements of the large number of variables included in the lagged RV measures of the Cross-Section, a similar intuition that a dynamic factor model provides. Our proposed framework provides robust forecasts of Realized Volatility  due to the advantage of our methodology in capturing the cross-sectional dependence which includes the time series evolution of risk aversion and economic sentiment reflected by the fluctuations in the Quadratic Variation of stock prices.     

Furthermore, the forecasting scheme (e.g., rolling versus recursive or expanding window) also changes the asymptotic distribution of the OOS distribution. In this paper, we focus on the rolling window forecasting scheme which ensures the validity of the forecast tests we consider for accessing the predictive accuracy of cross-sectional RVs. Secondly, another aspect for consideration is the forecast horizon we use, for example one-step vis-a-vis multiple forecasts require different treatment of the forecast evaluation tests, especially in the case of the use of the volatility proxy. In particular, the GW test in addition to the other two tests of unconditional predictive accuracy allows to verify whether the cross-sectional information set has predictive ability in forecasting RV. Since the GW test in constructed via a Wald type test using the loss differential based on the two competing models then we can compare the benchmark model and the forecast model in order to evaluate the null hypothesis of equal predictive accuracy. However, since we do not directly using the forecast error sequences but constructing the Wald statistic using the loss differential as the test function, then this allows to check the robustness of the Wald statistic to the chosen loss function as well as the model specification under the the null and alternative models. 
 
Specifically, the model specification (e.g., HAR, HARQ, HAR-J, CHAR) and the chosen shrinkage methodology (e.g, Standard Lasso, Adaptive Lasso, Elastic Net Lasso) can affect the RV forecasting. Firstly, the model specification captures stylized facts in the literature of stock Quadratic Variation which can affect the robustness and accuracy of forecasts. For example, the HARQ specification allows for measurement error correction  to the standard HAR model by including the RQ measure as an additional covariate. Thus, in the case of the Augmented HARQ model, this allows to investigate which cross-sectional measurement corrections of the cross-sections affect the RV forecasting of a particular firm. Using for example, the HAR-J model which includes the jump component of the Quadratic Variation provides another suitable model specification with particular aim to investigate which jumps of the cross-section affect the RV forecasting of a particular firm. Moreover, the shrinkage methodology allows to examine whether the effect of different degree of sparsity in the cross-sectional information set affect the RV forecasts using the forecast evaluation tests. 

In summary, the econometrician is interested to know the properties of the predictive accuracy tests under different economic conditions such as for nested versus non-nested environments especially for OOS applications. In this paper, we indeed examine the effects of the aforementioned features on the forecasting ability of the cross-section as these are reflected in the performance of both unconditional and conditional predictive ability. We are in particular interested to detect whether there are any specific patters for the firms of large cross-sections for which statistical evidence of non-equal predictive ability are supported by the majority of the predictive accuracy tests. The asymptotic properties of the predictive accuracy tests also depend on the assumptions regarding the uncertainty of the model parameters. When tests are constructed from forecast errors which are based on estimated rather than known parameters then this can affect the robustness of the tests especially for example in the case of unconditional predictive accuracy tests. 

\section{Conclusion}

Our research objective is to examine the predictive ability of a dynamic information set under cross-sectional dependence when forecasting realized volatility. Specifically, high-dimensional models under cross-section dependence have not seen much attention in the literature previously. In particular, examining the Cross-Section allows us to assess whether network effects can help improve high frequency volatility forecasts using lagged volatilities of other firms. In our paper, these network effects can be modelled as volatility spillover effects captured via the past volatilities of other firms. Our hypothesis of existence of predictability due to network effects of past volatilities induces via the Cross-Section of returns is indeed supported by related studies. \cite{chinco2019sparse} study the predictive accuracy of a Cross-Section of lagged returns for rolling one-minute-ahead return forecasts using the LASSO methodology. The LASSO appears to increase both out-of-sample fit and forecast-implied Sharpe ratios. The selection of covariates appears to be sparse and not based on a specific economic theory but on the statistical properties of LASSO in identifying a set of predictors. Similarly, in other related literature the authors identify industry effects, that is, the returns of larger stocks within an industry predict the future returns of smaller stocks within the same industry (see also \cite{hou2007industry}, \cite{ang2006cross} and \cite{cremers2015aggregate}). 

In this paper we investigate the Cross-Sectional firm dynamics as reflected by the unobservable quadratic variation of asset prices using as proxy the observable Realized Volatility measures of the Cross-Section. We focus on forecasting out-of-sample measures of risk such as Realized Volatility via the Cross-Section. Furthermore, we use forecast evaluation tests in order to access the predictive accuracy of the proposed methodology. We consider various econometric aspects related to the predictive accuracy of Cross-Sectional Realized Volatility such as the measurement error associated to the volatility proxy, the forecasting scheme and forecasting horizon as well as aspects related to the form of forecast comparisons. We focus on certain aspects related to the Cross-Sectional firm characteristics such as the industry specific characteristics in order to identify network volatility dynamics and spillover effects.

Our empirical study suggests that under the presence of cross-sectional dependence and multicollinearity the performance of traditional equal predictive accuracy tests can be affected due to the fact that Lasso shrinkage estimators suffer from over-shrinkage in high-dimensional problems, especially under the presence of multicollinearity. Strongly correlated data can cause over-shrinkage of the candidate predictors and thus when testing for equal predictive accuracy in high-dimensional models can suffer from over-rejections of the null hypothesis of equal predictive accuracy regardless of the presence of cross-sectional dependence. A novel equal predictive accuracy  procedure that corrects for such pitfalls will be useful for robust evaluation of forecasted error sequences regardless of cross-sectional dependence and the forecasting environment. A unified approach which bridge the gap for forecasts from non-nested, nested or overlapping high-dimensional models can be investigated in a future research paper.

\newpage

\section{Technical Results}

In this section, we provide proofs of related asymptotic theory which can be useful when constructing a robustified equal predictive accuracy testing procedure. To do this, we employ regularity conditions and theoretical results presented by \cite{kapetanios2018time}. Denote with $\beta^0$ the true population value of the standard Lasso estimator for simplicity. 

\medskip

\begin{theorem}
\label{Theorem1}
Let $y_t = x_t^{\prime} \beta_t^0 + \epsilon_t$ and $\hat{\beta}_t$ the time-varying Lasso estimator obtained as
\begin{align}\
\hat{\beta}_t 
= 
\underset{ \beta \in \mathbb{R}^{p} }{  \text{argmin} }  \ \left\{ \frac{1}{T_0} \sum_{t=1}^{ T } \left(  y_{j} -  x_{j}^{\prime} \beta \right)^2  +  \frac{ \lambda_{T_0}  }{ T_0 } \ \norm{ \beta }_1   \right\}
\end{align}
Then, it holds that $\hat{\beta}_t \overset{ p }{ \to } \beta_t^{0}, \ \text{as} \ \ T \to \infty$,
for all $t = [ \pi_0 T]$, $0 < \pi_0 < 1$, if $T_0 \to \infty$, $\left( \frac{ T_0 }{ T } \sqrt{ \text{log}(T_0)} \right) \to \infty$ as and $\lambda_{T_0} = \mathcal{O}(T_0)$.
\end{theorem}
\begin{proof}
For the proof of Theorem \ref{Theorem1} we can adapt the assumptions used by \cite{knight2000asymptotics}. In order to prove the required result, we need to show that 
\begin{align}
\label{pointwise.convergence}
\frac{1}{T_0}  \sum_{j=1}^T  w_{tj} \left( y_j - x_j^{\prime} \beta \right)^2 \overset{ p }{ \to } \left(  \beta_t^{0} - \beta \right)^{\prime} \Omega_T \left(  \beta_t^{0} - \beta \right) + \sigma^2 
\end{align}
where $T_0$ is the sample size of the out-of-sample period, such that $T_0 < T$. The pointwise convergence result given by expression \eqref{pointwise.convergence} holds for the information set $\mathcal{F}_{t}$. Note that pointwise convergence is sufficient for the asymptotic result we are after. We follow the proof given in Appendix B of \cite{kapetanios2018time}. We write
\begin{align}
\frac{1}{T_0} \sum_{j=1}^T w_{tj} \left( y_j - x_j^{\prime} \beta \right)^2 
&= \frac{1}{T_0} \sum_{j=1}^T w_{tj} \left[ x_j^{\prime} \left( \beta_j^{0} - \beta \right) \right]^2
+ \frac{2}{T_0} \sum_{j=1}^T w_{tj} \epsilon_j x_j^{\prime} \left( \beta_j^{0} - \beta \right)
\\
&+ \frac{1}{T_0} \sum_{j=1}^T w_{tj} \epsilon_j^2 
\equiv \sum_{j=1}^3 \mathcal{A}_{j} \left( T, T_0  \right) 
\end{align}
First, consider the expression  $\mathcal{A}_{1} \left( T, T_0  \right)$. We have that,  
\begin{align*}
\mathcal{A}_{1} \left( T, T_0  \right) \equiv \frac{1}{T_0} \sum_{j=1}^T w_{tj} \left[ x_j^{\prime} \left( \beta_j^{0} - \beta \right) \right]^2
&= 
\frac{3}{T_0} \sum_{j=1}^T w_{tj} \left[ x_j^{\prime} \left( \beta_j^{0} - \beta_t^{0} \right) \right]^2 
\\
&+ 
\left( \frac{2}{T_0} \sum_{j=1}^T w_{tj} x_j^{\prime} \left( \beta_j^{0} - \beta_t^{0} \right) x_j^{\prime} \right) \left( \beta_t^{0} - \beta \right) 
\\
&+  \left( \beta_t^{0} - \beta \right)^{\prime} \left( \frac{1}{T_0} \sum_{j=1}^{T_0} w_{tj} x_j x_j^{\prime} \right) \left( \beta_t^{0} - \beta \right)
\end{align*}

\newpage 

Assuming that the following convergence in probability holds
\begin{align}
\frac{1}{T_0} \sum_{j=1}^T w_{tj} x_j x_j^{\prime} \overset{ p }{\to} \Omega_T
\end{align}
Then, \cite{kapetanios2018time} shows that 
\begin{align}
\mathcal{A}_{1} \left( T, T_0  \right) \overset{ p }{\to}  \left( \beta_t^{0} - \beta \right)^{\prime}  \Omega_T  \left( \beta_t^{0} - \beta \right). 
\end{align}

Next, since $\mathbb{E} \left( x_t \epsilon_t \right) = 0$ and $\mathbb{E} \left( x_{i,t} \epsilon_t x_{j,t} \epsilon_s \right) = 0$, for all $i,j,s,t$ such that $t \neq s$, that is, no presence of serial correlation then this implies that $\mathbb{E} \left(  \mathcal{A}_{2} \left( T, T_0  \right)  \right) = 0$, and $\mathbb{E} \left(  \mathcal{A}^2_{2} \left( T, T_0  \right)  \right) \leq C \left( \sum_{j=1}^T w_{tj}^2 \right) \mathbb{E} ( \epsilon_k^2 ) \mathbb{E} \left( \sum_{i=1}^N x_{i,j}^2  \right) = \mathcal{O} \left( T_0^{-1} \right)$, since $\beta_T^{0}$ is bounded. Therefore, we can deduce that $\mathcal{A}_{2} \left( T, T_0  \right) = \mathcal{O} \left( T_0^{-1/2}  \right)$. Moreover, using the LLN it holds that  $\mathcal{A}_{3} \left( T, T_0  \right) = \sum_{i=1}^T  w_{tj} \epsilon_{j}^2 \overset{p}{\to}  \sigma^2$, which shows that expression \eqref{pointwise.convergence} holds. 
\end{proof} 

Next, we investigate another useful result for the development of the asymptotic theory of our framework by considering the time-varying lasso asymptotics as given by Theorem 2 in \cite{kapetanios2018time}.

\medskip

\begin{theorem}
\label{Theorem2}
Let $y_t = x_t \beta_t^0 + \epsilon_t$ and $\hat{\beta}$ the Lasso estimator obtained as
\begin{align}\
\hat{\beta}_t 
= 
\underset{ \beta \in \mathbb{R}^{p} }{  \text{argmin} }  \ \left\{ \frac{1}{T_0} \sum_{t=1}^{ T } \left(  y_{j} -  x_{j}^{\prime} \beta \right)^2  +  \lambda \norm{ \beta }_1  \right\}
\end{align}
where $\lambda \equiv \lambda_{T} = \text{log}( p_N )^2 T_0^{-1/2} + \displaystyle \frac{T_0}{T}$. Then, it holds that
\begin{align}
\frac{1}{T_0} \sum_{j=1}^T w_{tj} \left[ x_j^{\prime} \left( \beta_t^{0} - \hat{\beta}_t \right) \right]^2 \leq 3 \lambda \norm{ \beta_t^{0} }_1
\end{align}
with probability 1, as $p_N , T \to \infty$.
\end{theorem} 

\medskip

\begin{remark}
In particular, the limit result given by Theorem \ref{Theorem2}, provides a lower probability bound for the component of the Lasso optimization function which includes only the optimization function that corresponds to the standard OLS estimation procedure and can facilitate the development of relevant asymptotic results to our framework. Specifically, these bounds can be used to evaluate the convergence rates of the testing procedures. 
\end{remark}

\begin{proof}
Firstly, we notice that for a consistent Lasso estimator such that $\hat{\beta}_t \overset{ p }{ \to } \beta_t^{0}$, we obtain the following inequality
\begin{align}
\label{expression.to.proof}
\frac{1}{T_0} \sum_{j=1}^T w_{tj} \left( y_j - x_j^{\prime} \hat{\beta}_t  \right)^2 +  \lambda \norm{ \beta }_1 \leq \frac{1}{T_0} \sum_{j=1}^T w_{tj} \left( y_j - x_j^{\prime} \beta_t^0  \right)^2 +  \lambda \norm{ \beta_t^{0} }_1
\end{align}

\newpage 

Then, we have that 
\begin{align}
\label{expression1}
\frac{1}{T_0} \sum_{j=1}^T w_{tj} \left( y_j - x_j^{\prime} \hat{\beta}_t  \right)^2  
&= 
\frac{1}{T_0} \sum_{j=1}^T w_{tj} \left[ x_j^{\prime} \left( \beta_j^{0} - \beta_t^{0}  \right) \right]^2   
\nonumber
\\
&+ 
\frac{1}{T_0} \sum_{j=1}^T w_{tj} \left[ x_j^{\prime} \left( \beta_t^{0} - \hat{\beta}_t \right) \right]^2   \nonumber
\\
&+ 
\frac{2}{T_0} \sum_{j=1}^T w_{tj} \bigg[ x_j^{\prime} \left( \beta_j^{0} - \beta_t^{0} \right) \bigg]   \bigg[ x_j^{\prime} \left( \beta_t^{0} - \hat{\beta}_t \right) \bigg]
\nonumber
\\
&+
\frac{1}{T_0} \sum_{j=1}^T w_{tj} \epsilon_j^2 + \frac{1}{T_0}  \sum_{j=1}^T w_{tj} \epsilon_j x_j^{\prime} \left( \beta_j^{0} - \beta_t^{0} \right)
\nonumber
\\
&+ 
\frac{1}{T_0} \sum_{j=1}^T w_{tj} \epsilon_j x_j^{\prime} \left( \beta_t^{0} - \hat{ \beta }_t \right)
\end{align}

Then, consider the first term of the R.H.S of expression \eqref{expression.to.proof}, and expanding out we obtain the following expression 
\begin{align}
\label{expression2}
\frac{1}{T_0} \sum_{j=1}^T w_{tj} \left( y_j - x_j^{\prime} \beta_t^0  \right)^2 
&= 
\frac{1}{T_0} \sum_{j=1}^T w_{tj} \left[ x_j^{\prime} \left( \beta_j^0 - \beta_t^{0} \right) \right]
+ \frac{1}{T_0} \sum_{j=1}^T w_{tj} \epsilon_j^2  
\nonumber
\\
&-  \frac{1}{T_0} \sum_{j=1}^T w_{tj} \epsilon_j x_j^{\prime} \left( \beta_j^0  - \beta_t^{0}  \right) 
\end{align} 
Now, comparing expression \eqref{expression1} and expression \eqref{expression2}, by cancelling out terms and using the inequality given by expression \eqref{expression.to.proof}, we obtain the following
\begin{align}
& \left\{   
\frac{1}{T_0} \sum_{j=1}^T w_{tj} \left[ x_j^{\prime} \left( \beta_t^0 - \hat{\beta}_t^{0} \right) \right]^2 + \frac{2}{T_0} \sum_{j=1}^T w_{tj} \left[ x_j^{\prime} \left( \beta_j^0 - \beta_t^{0} \right)    x_j^{\prime} \left( \beta_t^0 - \hat{\beta}_t^{0} \right)  \right] + \lambda || \hat{\beta}_t ||_1
\right\}
\nonumber
\\
& \leq 
\left\{ \frac{1}{T_0} \sum_{j=1}^T w_{tj} \epsilon_j x_j^{\prime} \left( \beta_t^{0} - \hat{\beta}_t  \right) + \lambda \norm{ \beta_t^{0} }_1   \right\}
\end{align}
Therefore, after rearranging we obtain that 
\begin{align}
\frac{1}{T_0} \sum_{j=1}^T w_{tj} \left[ x_j^{\prime} \left( \beta_t^0 - \hat{\beta}_t^{0} \right) \right]^2  + \lambda \norm{ \hat{\beta}_t }_1 
&\leq \left( \frac{1}{T_0} \sum_{j=1}^T w_{tj} \epsilon_j x_j^{\prime} \right) \left( \hat{\beta}_t - \beta_t^{0} \right) 
\nonumber
\\
&- \left( \frac{2}{T_0} \sum_{j=1}^T w_{tj} \left[ x_j^{\prime} \left( \beta_j^{0} - \beta_t^{0}   \right)  x_j^{\prime} \right] \right) \left( \hat{\beta}_t - \beta_t^{0} \right) 
\nonumber
\\
&+ \lambda \norm{ \beta_t^{0} }_1
\end{align}   
Moreover, we have the following asymptotic results
\begin{align}
\frac{1}{T_0} \sum_{ | j - t|  > T_0 } w_{tj} \epsilon_j x_{i,j} = o_p (1) \ \ \ \text{and} \ \ \ \frac{2}{T_0} \sum_{ | j - t|  > T_0 } w_{tj} x_j^{\prime} \left( \beta_j^{0} - \beta_t^{0}   \right)  x_j^{\prime} = o_p (1)
\end{align}
Then, we obtain the expression below
\begin{align}
\label{inequality.expression}
& \left( \frac{1}{T_0} \sum_{ | j - t| \leq T_0 } w_{tj} \epsilon_j x_j^{\prime} \right) \left( \hat{\beta}_t - \beta_t^{0} \right) 
- \left( \frac{2}{T_0} \sum_{ | j - t| \leq T_0 } w_{tj} \bigg[ x_j^{\prime} \left( \beta_j^{0} - \beta_t^{0} \right)  x_j^{\prime} \bigg] \right) \left( \hat{\beta}_t - \beta_t^{0} \right) 
+ \lambda \norm{ \beta_t^{0} }_1
\nonumber
\\
& \leq 
\underset{ i \leq p_N }{ \text{max} } \left( \frac{1}{T_0} \left| \sum_{ | j - t| \leq T_0 } w_{tj} \epsilon_j x_{i,j} \right| \right) \norm{ \hat{\beta}_t - \beta_t^{0} }_1 + \underset{  | j - t| \leq T_0 }{ \text{max} } \norm{ x_j^{\prime} \left( \beta_j^{0} - \beta_t^{0} \right) } 
\nonumber
\\
&\times \left\{ \underset{ i \leq p_N }{ \text{max} } \left( \frac{2}{T_0} \sum_{ | j - t| \leq T_0 } w_{tj} \bigg[ \left| x_{i,j} \right| - \mathbb{E} \left( \left| x_{i,j} \right| \right) \bigg] \right)   +    \ \underset{ i \leq p_N }{ \text{max} } \left( \frac{2}{T_0} \sum_{ | j - t| \leq T_0 } w_{tj} \mathbb{E} \left( \left| x_{i,j} \right| \right) \right)  \right\} \times \norm{ \hat{\beta}_t^{0} - \beta_t^{0} }
\nonumber
\\
&+ \lambda \norm{ \beta_t^{0} }_1
\end{align}
Therefore, to prove the result of Theorem \ref{Theorem2} the following have to hold
\begin{align}
\label{show.a}
\underset{ i \leq p_N }{ \text{max} } \left( \frac{1}{T_0} \left| \sum_{ | j - t| \leq T_0 } w_{tj} \epsilon_j x_{i,j} \right| \right) \leq \lambda_0, 
\end{align} 
and that, 
\begin{align} 
\label{show.b}
\underset{ i \leq p_N }{ \text{max} } \left( \frac{2}{T_0} \sum_{ | j - t| \leq T_0 } w_{tj} \bigg[ \left| x_{i,j} \right| - \mathbb{E} \left( \left| x_{i,j} \right| \right) \bigg] \right)  \leq \lambda_0 
\end{align}
Detailed proofs of \eqref{show.a} and \eqref{show.b} can be found in \cite{kapetanios2018time} with an appropriate choice of $\lambda_T$ such that, $\lambda_T =  \left[ \text{log} \left( p_N  \right) \right]^2 T_0^{-1/2}$ and noting that
\begin{align}
\mathbb{P} \left(  \left| \sum_{ j = 1 }^T w_{tj} \epsilon_j x_{i,j} \right|  > \lambda_T \right) \leq \mathbb{P} \left(  \left| \sum_{ | j - t| \leq T_0 } w_{tj} \epsilon_j x_{i,j} \right|  > \frac{\lambda_T}{2} \right) + 
\mathbb{P} \left(  \left| \sum_{ | j - t| > T_0 } w_{tj} \epsilon_j x_{i,j} \right|  > \frac{\lambda_T}{2} \right)
\end{align}
Then, using \eqref{inequality.expression} and the asymptotic results $\underset{ i \leq p_N }{ \text{max} } \left( \frac{2}{T_0} \sum_{ | j - t| \leq T_0 } w_{tj} \mathbb{E} \left( \left| x_{i,j} \right| \right) \right) = \mathcal{O}(1)$, we obtain $\underset{ | j - t| \leq T_0 }{ \text{max} } \norm{ x_j^{\prime} \left( \beta_j^{0} - \beta_t^{0} \right) }   \times \underset{ i \leq p_N }{ \text{max} } \left( \frac{2}{T_0} \sum_{ | j - t| \leq T_0 } w_{tj} \mathbb{E} \left( \left| x_{i,j} \right| \right) \right) = \mathcal{O}(\frac{T_0}{T})$. Thus, by noting that $\norm{ \hat{\beta}_t - \beta_t^{0}}_1 \leq \norm{ \hat{\beta}_t }_1 + \norm{ \beta^{0}_t }_1$, and setting $\lambda \geq \lambda_0$, for $\lambda_0 = \left[ \text{log} \left( p_N \right) \right]^2 T_0^{-1/2} + \frac{T_0}{T}$, we prove the desirable result, that is, $\frac{1}{T_0} \sum_{j=1}^T w_{tj} \left[ x_j^{\prime} \left( \beta_t^{0} - \hat{\beta}_t \right) \right]^2 \leq 3 \lambda \norm{ \beta_t^{0} }_1$.
  
\end{proof}

\newpage 
      
\subsection{Estimation Results}  

\begin{small}
  
\begin{table}[htbp]
  \centering
  \caption{Summary statistics}
    \begin{tabular}{rrcccccc}
   \hline
    \multicolumn{1}{c}{\textbf{Company }} & \multicolumn{1}{c}{\textbf{Symbol}} & \textbf{Min} & \textbf{Mean} & \textbf{Median} & \textbf{Max} & \textbf{AR(1)} & \textbf{ARQ} \\
    \hline
    American Express & \multicolumn{1}{l}{AXP} & 0.088 & 4.603 & 2.184 & 290.338 & 0.602 & 0.9481 \\
    Boeing & \multicolumn{1}{l}{BA} & 0.167 & 3.371 & 2.147 & 79.76 & 0.630 & 0.822 \\
    Caterpillar & \multicolumn{1}{l}{CAT} & 0.207 & 3.81  & 2.401 & 127.119 & 0.727 & 0.8959 \\
    Cisco Systems & \multicolumn{1}{l}{CSCO} & 0.234 & 5.12  & 2.742 & 96.212 & 0.715 & 0.9411 \\
    Chevron & \multicolumn{1}{l}{CVX} & 0.105 & 2.286 & 1.483 & 139.984 & 0.653 & 1.0451 \\
    DuPont & \multicolumn{1}{l}{DD} & 0.093 & 3.327 & 2.165 & 81.721 & 0.707 & 0.9554 \\
    Walt Disney & \multicolumn{1}{l}{DIS} & 0.135 & 3.641 & 2.03  & 129.661 & 0.629 & 0.7719 \\
    General Electric & \multicolumn{1}{l}{GE} & 0.131 & 3.44  & 1.794 & 173.223 & 0.681 & 0.9866 \\
    The Home Deport  & \multicolumn{1}{l}{HD} & 0.171 & 3.798 & 2.161 & 133.855 & 0.633 & 0.9916 \\
    IBM   & \multicolumn{1}{l}{IBM} & 0.115 & 2.464 & 1.34  & 72.789 & 0.654 & 0.8899 \\
    Intel & \multicolumn{1}{l}{INTC} & 0.208 & 4.654 & 2.674 & 89.735 & 0.731 & 0.9678 \\
    Johnson \& Johnson & \multicolumn{1}{l}{JNJ} & 0.062 & 1.68  & 0.999 & 58.338 & 0.614 & 0.9329 \\
    JPMorgan Chase & \multicolumn{1}{l}{JPM} & 0.114 & 5.42  & 2.552 & 261.459 & 0.716 & 1.0603 \\
    Coca-Cola & \multicolumn{1}{l}{KO} & 0.049 & 2.011 & 1.154 & 54.883 & 0.618 & 0.8336 \\
    McDonalds's & \multicolumn{1}{l}{MCD} & 0.09  & 2.678 & 1.68  & 130.103 & 0.390 & 0.6724 \\
    3M    & \multicolumn{1}{l}{MMM} & 0.14  & 2.278 & 1.358 & 123.197 & 0.495 & 0.7474 \\
    Merck & \multicolumn{1}{l}{MRK} & 0.127 & 2.758 & 1.718 & 223.723 & 0.372 & 0.7074 \\
    Microsoft & \multicolumn{1}{l}{MSFT} & 0.166 & 3.087 & 1.824 & 59.164 & 0.718 & 0.8891 \\
    Nike  & \multicolumn{1}{l}{NKE} & 0.192 & 3.431 & 1.98  & 84.338 & 0.580 & 0.7833 \\
    Pfizer & \multicolumn{1}{l}{PFE} & 0.176 & 2.822 & 1.809 & 60.302 & 0.570 & 0.8371 \\
    Procter \& Gamble & \multicolumn{1}{l}{PG} & 0.085 & 2.007 & 1.064 & 80.124 & 0.587 & 0.7853 \\
    Travelers & \multicolumn{1}{l}{TRV} & 0.098 & 3.579 & 1.637 & 273.579 & 0.647 & 0.9149 \\
    UnitedHealth Group & \multicolumn{1}{l}{UNH} & 0.222 & 4.145 & 2.304 & 169.815 & 0.614 & 0.8436 \\
    United Technologies & \multicolumn{1}{l}{UTX} & 0.126 & 2.793 & 1.658 & 92.105 & 0.648 & 0.8821 \\
    Verizon & \multicolumn{1}{l}{VZ} & 0.145 & 2.788 & 1.637 & 99.821 & 0.646 & 0.8579 \\
    Walt-Mart & \multicolumn{1}{l}{WMT} & 0.148 & 2.761 & 1.443 & 114.639 & 0.611 & 0.8098 \\
    ExxonMobil & \multicolumn{1}{l}{XOM} & 0.114 & 2.348 & 1.476 & 130.667 & 0.668 & 0.9958 \\
     \hline
    \end{tabular}%
  \label{Table1}%
\end{table}%

\begin{table}[h!]
  \centering
     \begin{tabular}{llrrrrr}
    \hline
    \textbf{Benchmark Model } & Average Measure & \multicolumn{1}{c}{HAR} & \multicolumn{1}{c}{HARQ} & \multicolumn{1}{c}{HARQ-F} & \multicolumn{1}{c}{HAR-J } & \multicolumn{1}{c}{CHAR} \\
    \hline
    \multirow{2}[0]{*}{OLS estimation} & $R^2$    &  0.485     &  0.509   &   0.514  &  0.491  &  0.489 \\
          & MSE   & 14.985 &  14.170   &  14.005  &  14.822  & 14.927 
\\
    \hline
    \end{tabular}%
  \label{table1a2}%
\end{table}%

\end{small}

Table \ref{Table1} (see also Table 2 from BPQ) provides summary statistics for the daily RV for each of the 27 firms of DJIA from the dataset of \cite{bollerslev2016exploiting}. The full sample size corresponds to 4202 daily RV observations across all firms from April 21, 1997 to December 31, 2013. The column AR reports the standard first order autocorrelation coefficient and the column ARQ refers to the $\beta_1$ estimates of the ARQ model.

\newpage 
    
\begin{small}

\begin{table}[htbp]
  \centering
  \caption{Out-of-sample estimation results (Rolling Window P = 252)}
    \begin{tabular}{rrrrrrr}
     \hline
          & \multicolumn{6}{c}{Diebold \& Mariano Test} \\
    \hline
          & \multicolumn{1}{l}{Model 1:} & \multicolumn{1}{l}{Standard HAR} &       & Standard HAR &       & Standard HAR \\
          & \multicolumn{1}{l}{Model 2:} & \multicolumn{1}{l}{Lasso} &       & Adaptive Lasso &       & Elastic Lasso \\
           \hline
    \multicolumn{1}{c}{\textbf{}} & \multicolumn{1}{c}{\textbf{DM Test}} & \multicolumn{1}{c}{\textbf{p-value}} & \multicolumn{1}{c}{\textbf{DM Test}} & \multicolumn{1}{c}{\textbf{p-value}} & \multicolumn{1}{c}{\textbf{DM Test}} & \multicolumn{1}{c}{\textbf{p-value}} \\
     \hline
    \multicolumn{1}{c}{\textbf{1}} & \multicolumn{1}{c}{-6.515} & \multicolumn{1}{c}{\textbf{0.000}} & \multicolumn{1}{c}{-1.652} & \multicolumn{1}{c}{0.098} & \multicolumn{1}{c}{0.953} & \multicolumn{1}{c}{0.340} \\
    \multicolumn{1}{c}{\textbf{2}} & \multicolumn{1}{c}{1.611} & \multicolumn{1}{c}{0.107} & \multicolumn{1}{c}{1.661} & \multicolumn{1}{c}{0.097} & \multicolumn{1}{c}{1.434} & \multicolumn{1}{c}{0.151} \\
    \multicolumn{1}{c}{\textbf{3}} & \multicolumn{1}{c}{-9.198} & \multicolumn{1}{c}{\textbf{0.000}} & \multicolumn{1}{c}{-6.124} & \multicolumn{1}{c}{\textbf{0.000}} & \multicolumn{1}{c}{1.730} & \multicolumn{1}{c}{0.084} \\
    \multicolumn{1}{c}{\textbf{4}} & \multicolumn{1}{c}{-2.106} & \multicolumn{1}{c}{\textbf{0.035}} & \multicolumn{1}{c}{-2.375} & \multicolumn{1}{c}{\textbf{0.018}} & \multicolumn{1}{c}{0.945} & \multicolumn{1}{c}{0.345} \\
    \multicolumn{1}{c}{\textbf{5}} & \multicolumn{1}{c}{0.771} & \multicolumn{1}{c}{0.441} & \multicolumn{1}{c}{0.151} & \multicolumn{1}{c}{0.880} & \multicolumn{1}{c}{0.600} & \multicolumn{1}{c}{0.548} \\
    \multicolumn{1}{c}{\textbf{6}} & \multicolumn{1}{c}{-4.474} & \multicolumn{1}{c}{\textbf{0.000}} & \multicolumn{1}{c}{-1.302} & \multicolumn{1}{c}{0.193} & \multicolumn{1}{c}{1.957} & \multicolumn{1}{c}{\textbf{0.050}} \\
    \multicolumn{1}{c}{\textbf{7}} & \multicolumn{1}{c}{-2.370} & \multicolumn{1}{c}{\textbf{0.018}} & \multicolumn{1}{c}{-1.986} & \multicolumn{1}{c}{0.047} & \multicolumn{1}{c}{0.941} & \multicolumn{1}{c}{0.346} \\
    \multicolumn{1}{c}{\textbf{8}} & \multicolumn{1}{c}{-7.673} & \multicolumn{1}{c}{\textbf{0.000}} & \multicolumn{1}{c}{-5.942} & \multicolumn{1}{c}{\textbf{0.000}} & \multicolumn{1}{c}{4.238} & \multicolumn{1}{c}{\textbf{0.000}} \\
    \multicolumn{1}{c}{\textbf{9}} & \multicolumn{1}{c}{-3.269} & \multicolumn{1}{c}{\textbf{0.001}} & \multicolumn{1}{c}{-2.255} & \multicolumn{1}{c}{\textbf{0.024}} & \multicolumn{1}{c}{0.498} & \multicolumn{1}{c}{0.618} \\
    \multicolumn{1}{c}{\textbf{10}} & \multicolumn{1}{c}{1.165} & \multicolumn{1}{c}{0.244} & \multicolumn{1}{c}{1.085} & \multicolumn{1}{c}{0.278} & \multicolumn{1}{c}{1.998} & \multicolumn{1}{c}{\textbf{0.046}} \\
    \multicolumn{1}{c}{\textbf{11}} & \multicolumn{1}{c}{-6.135} & \multicolumn{1}{c}{\textbf{0.000}} & \multicolumn{1}{c}{-3.544} & \multicolumn{1}{c}{\textbf{0.000}} & \multicolumn{1}{c}{2.668} & \multicolumn{1}{c}{\textbf{0.008}} \\
    \multicolumn{1}{c}{\textbf{12}} & \multicolumn{1}{c}{0.881} & \multicolumn{1}{c}{0.378} & \multicolumn{1}{c}{-1.913} & \multicolumn{1}{c}{0.056} & \multicolumn{1}{c}{2.629} & \multicolumn{1}{c}{\textbf{0.009}} \\
    \multicolumn{1}{c}{\textbf{13}} & \multicolumn{1}{c}{-3.026} & \multicolumn{1}{c}{\textbf{0.002}} & \multicolumn{1}{c}{2.384} & \multicolumn{1}{c}{\textbf{0.017}} & \multicolumn{1}{c}{0.753} & \multicolumn{1}{c}{0.451} \\
    \multicolumn{1}{c}{\textbf{14}} & \multicolumn{1}{c}{-1.132} & \multicolumn{1}{c}{0.257} & \multicolumn{1}{c}{1.214} & \multicolumn{1}{c}{0.225} & \multicolumn{1}{c}{-0.601} & \multicolumn{1}{c}{0.548} \\
    \multicolumn{1}{c}{\textbf{15}} & \multicolumn{1}{c}{-18.677} & \multicolumn{1}{c}{\textbf{0.000}} & \multicolumn{1}{c}{-14.626} & \multicolumn{1}{c}{\textbf{0.000}} & \multicolumn{1}{c}{9.164} & \multicolumn{1}{c}{\textbf{0.000}} \\
    \multicolumn{1}{c}{\textbf{16}} & \multicolumn{1}{c}{-2.123} & \multicolumn{1}{c}{\textbf{0.034}} & \multicolumn{1}{c}{-0.536} & \multicolumn{1}{c}{0.592} & \multicolumn{1}{c}{4.143} & \multicolumn{1}{c}{\textbf{0.000}} \\
    \multicolumn{1}{c}{\textbf{17}} & \multicolumn{1}{c}{-5.258} & \multicolumn{1}{c}{\textbf{0.000}} & \multicolumn{1}{c}{-6.301} & \multicolumn{1}{c}{\textbf{0.000}} & \multicolumn{1}{c}{1.645} & \multicolumn{1}{c}{0.100} \\
    \multicolumn{1}{c}{\textbf{18}} & \multicolumn{1}{c}{2.473} & \multicolumn{1}{c}{\textbf{0.013}} & \multicolumn{1}{c}{0.347} & \multicolumn{1}{c}{0.729} & \multicolumn{1}{c}{0.685} & \multicolumn{1}{c}{0.494} \\
    \multicolumn{1}{c}{\textbf{19}} & \multicolumn{1}{c}{-2.282} & \multicolumn{1}{c}{\textbf{0.023}} & \multicolumn{1}{c}{1.050} & \multicolumn{1}{c}{0.294} & \multicolumn{1}{c}{1.794} & \multicolumn{1}{c}{0.073} \\
    \multicolumn{1}{c}{\textbf{20}} & \multicolumn{1}{c}{-6.107} & \multicolumn{1}{c}{\textbf{0.000}} & \multicolumn{1}{c}{-2.493} & \multicolumn{1}{c}{\textbf{0.013}} & \multicolumn{1}{c}{0.298} & \multicolumn{1}{c}{0.765} \\
    \multicolumn{1}{c}{\textbf{21}} & \multicolumn{1}{c}{2.706} & \multicolumn{1}{c}{\textbf{0.007}} & \multicolumn{1}{c}{3.851} & \multicolumn{1}{c}{\textbf{0.000}} & \multicolumn{1}{c}{2.060} & \multicolumn{1}{c}{\textbf{0.039}} \\
    \multicolumn{1}{c}{\textbf{22}} & \multicolumn{1}{c}{1.093} & \multicolumn{1}{c}{0.274} & \multicolumn{1}{c}{2.826} & \multicolumn{1}{c}{\textbf{0.005}} & \multicolumn{1}{c}{2.462} & \multicolumn{1}{c}{\textbf{0.014}} \\
    \multicolumn{1}{c}{\textbf{23}} & \multicolumn{1}{c}{-0.810} & \multicolumn{1}{c}{0.418} & \multicolumn{1}{c}{-0.689} & \multicolumn{1}{c}{0.491} & \multicolumn{1}{c}{0.247} & \multicolumn{1}{c}{0.805} \\
    \multicolumn{1}{c}{\textbf{24}} & \multicolumn{1}{c}{-0.491} & \multicolumn{1}{c}{0.623} & \multicolumn{1}{c}{-0.261} & \multicolumn{1}{c}{0.794} & \multicolumn{1}{c}{1.301} & \multicolumn{1}{c}{0.193} \\
    \multicolumn{1}{c}{\textbf{25}} & \multicolumn{1}{c}{-0.445} & \multicolumn{1}{c}{0.657} & \multicolumn{1}{c}{1.693} & \multicolumn{1}{c}{0.090} & \multicolumn{1}{c}{1.736} & \multicolumn{1}{c}{0.083} \\
    \multicolumn{1}{c}{\textbf{26}} & \multicolumn{1}{c}{-5.640} & \multicolumn{1}{c}{\textbf{0.000}} & \multicolumn{1}{c}{-1.607} & \multicolumn{1}{c}{0.108} & \multicolumn{1}{c}{-0.048} & \multicolumn{1}{c}{0.962} \\
    \multicolumn{1}{c}{\textbf{27}} & \multicolumn{1}{c}{-5.602} & \multicolumn{1}{c}{\textbf{0.000}} & \multicolumn{1}{c}{-3.587} & \multicolumn{1}{c}{\textbf{0.000}} & \multicolumn{1}{c}{1.889} & \multicolumn{1}{c}{0.059} \\
  \hline
    \end{tabular}%
  \label{Table2}%
\end{table}%

\end{small}

Table \ref{Table2} demonstrates the OOS Diebold and Mariano test statistics with corresponding p-values to indicate the statistical significance of the cross-sectional predictability of Realized Volatility across the 27 firms from April 21, 1997 to December 31, 2013. We consider a rolling window forecasting scheme with window size $T_0 = 252$. Moreover, the forecast sequences are based on Model 1 corresponds to the standard HAR model and Model 2 which corresponds to the forecast model with model specifications being the the standard Lasso, Adaptive Lasso and Elastic Lasso.

\newpage 

\begin{small}

\begin{table}[htbp]
  \centering
  \caption{Out-of-sample estimation results (Rolling Window P = 252)}
    \begin{tabular}{rrrrrrr}
   \hline
          & \multicolumn{6}{c}{\textbf{Clark \& West Test}} \\
    \hline
          & \multicolumn{1}{l}{Model 1:} & \multicolumn{1}{l}{Standard HAR} &       & Standard HAR &       & Standard HAR \\
          & \multicolumn{1}{l}{Model 2:} & \multicolumn{1}{l}{Lasso} &       & Adaptive Lasso &       & Elastic Lasso \\
           \hline
    \multicolumn{1}{c}{\textbf{}} & \multicolumn{1}{c}{\textbf{CW Test}} & \multicolumn{1}{c}{\textbf{p-value}} & \multicolumn{1}{c}{\textbf{CW Test}} & \multicolumn{1}{c}{\textbf{p-value}} & \multicolumn{1}{c}{\textbf{CW Test}} & \multicolumn{1}{c}{\textbf{p-value}} \\
     \hline
    \multicolumn{1}{c}{\textbf{1}} & \multicolumn{1}{c}{-4.710} & \multicolumn{1}{c}{1.000} & \multicolumn{1}{c}{-0.236} & \multicolumn{1}{c}{0.593} & \multicolumn{1}{c}{3.388} & \multicolumn{1}{c}{\textbf{0.000}} \\
    \multicolumn{1}{c}{\textbf{2}} & \multicolumn{1}{c}{1.696} & \multicolumn{1}{c}{\textbf{0.046}} & \multicolumn{1}{c}{1.726} & \multicolumn{1}{c}{\textbf{0.043}} & \multicolumn{1}{c}{1.898} & \multicolumn{1}{c}{\textbf{0.029}} \\
    \multicolumn{1}{c}{\textbf{3}} & \multicolumn{1}{c}{-4.463} & \multicolumn{1}{c}{1.000} & \multicolumn{1}{c}{-3.688} & \multicolumn{1}{c}{1.000} & \multicolumn{1}{c}{4.855} & \multicolumn{1}{c}{\textbf{0.000}} \\
    \multicolumn{1}{c}{\textbf{4}} & \multicolumn{1}{c}{-0.487} & \multicolumn{1}{c}{0.687} & \multicolumn{1}{c}{-0.362} & \multicolumn{1}{c}{0.641} & \multicolumn{1}{c}{2.997} & \multicolumn{1}{c}{\textbf{0.002}} \\
    \multicolumn{1}{c}{\textbf{5}} & \multicolumn{1}{c}{1.987} & \multicolumn{1}{c}{\textbf{0.024}} & \multicolumn{1}{c}{1.329} & \multicolumn{1}{c}{0.093} & \multicolumn{1}{c}{7.958} & \multicolumn{1}{c}{\textbf{0.000}} \\
    \multicolumn{1}{c}{\textbf{6}} & \multicolumn{1}{c}{-1.689} & \multicolumn{1}{c}{0.954} & \multicolumn{1}{c}{-0.130} & \multicolumn{1}{c}{0.552} & \multicolumn{1}{c}{3.178} & \multicolumn{1}{c}{\textbf{0.001}} \\
    \multicolumn{1}{c}{\textbf{7}} & \multicolumn{1}{c}{-0.951} & \multicolumn{1}{c}{0.829} & \multicolumn{1}{c}{-0.830} & \multicolumn{1}{c}{0.796} & \multicolumn{1}{c}{1.934} & \multicolumn{1}{c}{\textbf{0.027}} \\
    \multicolumn{1}{c}{\textbf{8}} & \multicolumn{1}{c}{-6.019} & \multicolumn{1}{c}{1.000} & \multicolumn{1}{c}{-4.261} & \multicolumn{1}{c}{1.000} & \multicolumn{1}{c}{10.687} & \multicolumn{1}{c}{\textbf{0.000}} \\
    \multicolumn{1}{c}{\textbf{9}} & \multicolumn{1}{c}{-1.237} & \multicolumn{1}{c}{0.891} & \multicolumn{1}{c}{-0.341} & \multicolumn{1}{c}{0.633} & \multicolumn{1}{c}{3.201} & \multicolumn{1}{c}{\textbf{0.001}} \\
    \multicolumn{1}{c}{\textbf{10}} & \multicolumn{1}{c}{2.555} & \multicolumn{1}{c}{\textbf{0.006}} & \multicolumn{1}{c}{2.519} & \multicolumn{1}{c}{\textbf{0.006}} & \multicolumn{1}{c}{3.997} & \multicolumn{1}{c}{\textbf{0.000}} \\
    \multicolumn{1}{c}{\textbf{11}} & \multicolumn{1}{c}{-3.390} & \multicolumn{1}{c}{1.000} & \multicolumn{1}{c}{-2.130} & \multicolumn{1}{c}{0.983} & \multicolumn{1}{c}{5.191} & \multicolumn{1}{c}{\textbf{0.000}} \\
    \multicolumn{1}{c}{\textbf{12}} & \multicolumn{1}{c}{2.283} & \multicolumn{1}{c}{\textbf{0.012}} & \multicolumn{1}{c}{-0.280} & \multicolumn{1}{c}{0.610} & \multicolumn{1}{c}{5.429} & \multicolumn{1}{c}{\textbf{0.000}} \\
    \multicolumn{1}{c}{\textbf{13}} & \multicolumn{1}{c}{-1.817} & \multicolumn{1}{c}{0.965} & \multicolumn{1}{c}{5.146} & \multicolumn{1}{c}{\textbf{0.000}} & \multicolumn{1}{c}{3.321} & \multicolumn{1}{c}{\textbf{0.001}} \\
    \multicolumn{1}{c}{\textbf{14}} & \multicolumn{1}{c}{0.919} & \multicolumn{1}{c}{0.179} & \multicolumn{1}{c}{2.397} & \multicolumn{1}{c}{\textbf{0.009}} & \multicolumn{1}{c}{4.023} & \multicolumn{1}{c}{\textbf{0.000}} \\
    \multicolumn{1}{c}{\textbf{15}} & \multicolumn{1}{c}{-14.866} & \multicolumn{1}{c}{1.000} & \multicolumn{1}{c}{-11.890} & \multicolumn{1}{c}{1.000} & \multicolumn{1}{c}{14.476} & \multicolumn{1}{c}{\textbf{0.000}} \\
    \multicolumn{1}{c}{\textbf{16}} & \multicolumn{1}{c}{-0.515} & \multicolumn{1}{c}{0.697} & \multicolumn{1}{c}{0.510} & \multicolumn{1}{c}{0.305} & \multicolumn{1}{c}{6.891} & \multicolumn{1}{c}{\textbf{0.000}} \\
    \multicolumn{1}{c}{\textbf{17}} & \multicolumn{1}{c}{-3.081} & \multicolumn{1}{c}{0.999} & \multicolumn{1}{c}{-4.046} & \multicolumn{1}{c}{1.000} & \multicolumn{1}{c}{10.691} & \multicolumn{1}{c}{\textbf{0.000}} \\
    \multicolumn{1}{c}{\textbf{18}} & \multicolumn{1}{c}{3.140} & \multicolumn{1}{c}{\textbf{0.001}} & \multicolumn{1}{c}{1.136} & \multicolumn{1}{c}{0.129} & \multicolumn{1}{c}{3.733} & \multicolumn{1}{c}{\textbf{0.000}} \\
    \multicolumn{1}{c}{\textbf{19}} & \multicolumn{1}{c}{-0.657} & \multicolumn{1}{c}{0.744} & \multicolumn{1}{c}{2.222} & \multicolumn{1}{c}{0.014} & \multicolumn{1}{c}{5.276} & \multicolumn{1}{c}{\textbf{0.000}} \\
    \multicolumn{1}{c}{\textbf{20}} & \multicolumn{1}{c}{-2.738} & \multicolumn{1}{c}{0.997} & \multicolumn{1}{c}{-0.894} & \multicolumn{1}{c}{0.814} & \multicolumn{1}{c}{4.012} & \multicolumn{1}{c}{\textbf{0.000}} \\
    \multicolumn{1}{c}{\textbf{21}} & \multicolumn{1}{c}{3.548} & \multicolumn{1}{c}{\textbf{0.000}} & \multicolumn{1}{c}{5.408} & \multicolumn{1}{c}{\textbf{0.000}} & \multicolumn{1}{c}{4.694} & \multicolumn{1}{c}{\textbf{0.000}} \\
    \multicolumn{1}{c}{\textbf{22}} & \multicolumn{1}{c}{2.054} & \multicolumn{1}{c}{\textbf{0.021}} & \multicolumn{1}{c}{3.632} & \multicolumn{1}{c}{\textbf{0.000}} & \multicolumn{1}{c}{4.516} & \multicolumn{1}{c}{\textbf{0.000}} \\
    \multicolumn{1}{c}{\textbf{23}} & \multicolumn{1}{c}{0.304} & \multicolumn{1}{c}{0.381} & \multicolumn{1}{c}{0.242} & \multicolumn{1}{c}{0.404} & \multicolumn{1}{c}{3.630} & \multicolumn{1}{c}{\textbf{0.000}} \\
    \multicolumn{1}{c}{\textbf{24}} & \multicolumn{1}{c}{1.229} & \multicolumn{1}{c}{0.110} & \multicolumn{1}{c}{1.444} & \multicolumn{1}{c}{0.075} & \multicolumn{1}{c}{4.773} & \multicolumn{1}{c}{\textbf{0.000}} \\
    \multicolumn{1}{c}{\textbf{25}} & \multicolumn{1}{c}{0.839} & \multicolumn{1}{c}{0.201} & \multicolumn{1}{c}{2.728} & \multicolumn{1}{c}{\textbf{0.003}} & \multicolumn{1}{c}{4.389} & \multicolumn{1}{c}{\textbf{0.000}} \\
    \multicolumn{1}{c}{\textbf{26}} & \multicolumn{1}{c}{-3.544} & \multicolumn{1}{c}{1.000} & \multicolumn{1}{c}{-0.635} & \multicolumn{1}{c}{0.737} & \multicolumn{1}{c}{1.885} & \multicolumn{1}{c}{\textbf{0.030}} \\
    \multicolumn{1}{c}{\textbf{27}} & \multicolumn{1}{c}{-4.168} & \multicolumn{1}{c}{1.000} & \multicolumn{1}{c}{-2.102} & \multicolumn{1}{c}{0.982} & \multicolumn{1}{c}{6.919} & \multicolumn{1}{c}{\textbf{0.000}} \\
    \hline
    \end{tabular}%
  \label{Table3}%
\end{table}%

\end{small}

Table \ref{Table3} demonstrates the OOS Clark and West test statistics with corresponding p-values to indicate the statistical significance of the cross-sectional predictability of Realized Volatility across the 27 firms from April 21, 1997 to December 31, 2013. We consider a rolling window forecasting scheme with window size $T_0 = 252$. Moreover, the forecast sequences are based on Model 1 corresponds to the standard HAR model and Model 2 which corresponds to the forecast model with model specifications being the the standard Lasso, Adaptive Lasso and Elastic Lasso.

\newpage

\bibliographystyle{apalike}

{\small
\bibliography{myreferences2}}

\begin{thebibliography}{}

\bibitem[A{\"\i}t-Sahalia and Jacod, 2014]{ait2014high}
A{\"\i}t-Sahalia, Y. and Jacod, J. (2014).
\newblock {\em High-frequency financial econometrics}.
\newblock Princeton University Press.

\bibitem[Andersen and Bollerslev, 1998]{andersen1998answering}
Andersen, T.~G. and Bollerslev, T. (1998).
\newblock Answering the skeptics: Yes, standard volatility models do provide
  accurate forecasts.
\newblock {\em International economic review}, pages 885--905.

\bibitem[Andersen et~al., 2007]{andersen2007roughing}
Andersen, T.~G., Bollerslev, T., and Diebold, F.~X. (2007).
\newblock Roughing it up: Including jump components in the measurement,
  modeling, and forecasting of return volatility.
\newblock {\em The review of economics and statistics}, 89(4):701--720.

\bibitem[Andersen et~al., 2001]{andersen2001distribution}
Andersen, T.~G., Bollerslev, T., Diebold, F.~X., and Ebens, H. (2001).
\newblock The distribution of realized stock return volatility.
\newblock {\em Journal of financial economics}, 61(1):43--76.

\bibitem[Andersen et~al., 2003]{andersen2003modeling}
Andersen, T.~G., Bollerslev, T., Diebold, F.~X., and Labys, P. (2003).
\newblock Modeling and forecasting realized volatility.
\newblock {\em Econometrica}, 71(2):579--625.

\bibitem[Andreou, 2016]{andreou2016use}
Andreou, E. (2016).
\newblock On the use of high frequency measures of volatility in midas
  regressions.
\newblock {\em Journal of Econometrics}, 193(2):367--389.

\bibitem[Andreou et~al., 2019]{andreou2019inference}
Andreou, E., Gagliardini, P., Ghysels, E., and Rubin, M. (2019).
\newblock Inference in group factor models with an application to
  mixed-frequency data.
\newblock {\em Econometrica}, 87(4):1267--1305.

\bibitem[Andreou et~al., 2010]{andreou2010regression}
Andreou, E., Ghysels, E., and Kourtellos, A. (2010).
\newblock Regression models with mixed sampling frequencies.
\newblock {\em Journal of Econometrics}, 158(2):246--261.

\bibitem[Andrews, 2005]{andrews2005cross}
Andrews, D.~W. (2005).
\newblock Cross-section regression with common shocks.
\newblock {\em Econometrica}, 73(5):1551--1585.

\bibitem[Ang et~al., 2006]{ang2006cross}
Ang, A., Hodrick, R.~J., Xing, Y., and Zhang, X. (2006).
\newblock The cross-section of volatility and expected returns.
\newblock {\em The Journal of Finance}, 61(1):259--299.

\bibitem[Audrino and Knaus, 2016]{audrino2016lassoing}
Audrino, F. and Knaus, S.~D. (2016).
\newblock Lassoing the har model: A model selection perspective on realized
  volatility dynamics.
\newblock {\em Econometric Reviews}, 35(8-10):1485--1521.

\bibitem[Baker et~al., 2016]{baker2016measuring}
Baker, S.~R., Bloom, N., and Davis, S.~J. (2016).
\newblock Measuring economic policy uncertainty.
\newblock {\em The quarterly journal of economics}, 131(4):1593--1636.

\bibitem[Balestra and Nerlove, 1966]{balestra1966pooling}
Balestra, P. and Nerlove, M. (1966).
\newblock Pooling cross section and time series data in the estimation of a
  dynamic model: The demand for natural gas.
\newblock {\em Econometrica: Journal of the econometric society}, pages
  585--612.

\bibitem[Barndorff-Nielsen et~al., 2008a]{barndorff2008designing}
Barndorff-Nielsen, O.~E., Hansen, P.~R., Lunde, A., and Shephard, N. (2008a).
\newblock Designing realized kernels to measure the ex post variation of equity
  prices in the presence of noise.
\newblock {\em Econometrica}, 76(6):1481--1536.

\bibitem[Barndorff-Nielsen et~al., 2008b]{barndorff2008measuring}
Barndorff-Nielsen, O.~E., Kinnebrock, S., and Shephard, N. (2008b).
\newblock Measuring downside risk-realised semivariance.
\newblock {\em CREATES Research Paper}, (2008-42).

\bibitem[Barndorff-Nielsen and Shephard, 2002]{barndorff2002econometric}
Barndorff-Nielsen, O.~E. and Shephard, N. (2002).
\newblock Econometric analysis of realized volatility and its use in estimating
  stochastic volatility models.
\newblock {\em Journal of the Royal Statistical Society: Series B (Statistical
  Methodology)}, 64(2):253--280.

\bibitem[Barndorff-Nielsen and Shephard, 2004]{barndorff2004power}
Barndorff-Nielsen, O.~E. and Shephard, N. (2004).
\newblock Power and bipower variation with stochastic volatility and jumps.
\newblock {\em Journal of financial econometrics}, 2(1):1--37.

\bibitem[Billio et~al., 2012]{billio2012econometric}
Billio, M., Getmansky, M., Lo, A.~W., and Pelizzon, L. (2012).
\newblock Econometric measures of connectedness and systemic risk in the
  finance and insurance sectors.
\newblock {\em Journal of financial economics}, 104(3):535--559.

\bibitem[Bollerslev, 1986]{bollerslev1986generalized}
Bollerslev, T. (1986).
\newblock Generalized autoregressive conditional heteroskedasticity.
\newblock {\em Journal of econometrics}, 31(3):307--327.

\bibitem[Bollerslev et~al., 2018]{bollerslev2018risk}
Bollerslev, T., Hood, B., Huss, J., and Pedersen, L.~H. (2018).
\newblock Risk everywhere: Modeling and managing volatility.
\newblock {\em The Review of Financial Studies}, 31(7):2729--2773.

\bibitem[Bollerslev et~al., 2020]{bollerslev2020realized}
Bollerslev, T., Li, J., Patton, A.~J., and Quaedvlieg, R. (2020).
\newblock Realized semicovariances.
\newblock {\em Econometrica}, 88(4):1515--1551.

\bibitem[Bollerslev et~al., 2016]{bollerslev2016exploiting}
Bollerslev, T., Patton, A.~J., and Quaedvlieg, R. (2016).
\newblock Exploiting the errors: A simple approach for improved volatility
  forecasting.
\newblock {\em Journal of Econometrics}, 192(1):1--18.

\bibitem[Busetti and Marcucci, 2013]{busetti2013comparing}
Busetti, F. and Marcucci, J. (2013).
\newblock Comparing forecast accuracy: a monte carlo investigation.
\newblock {\em International Journal of Forecasting}, 29(1):13--27.

\bibitem[Chetverikov et~al., 2020]{chetverikov2020cross}
Chetverikov, D., Liao, Z., and Chernozhukov, V. (2020).
\newblock On cross-validated lasso in high dimensions.
\newblock {\em The Annals of Statistics}, 48(5):1--54.

\bibitem[Chinco et~al., 2019]{chinco2019sparse}
Chinco, A.~M., Clark-Joseph, A.~D., and Ye, M. (2019).
\newblock Sparse signals in the cross-section of returns.
\newblock {\em The Journal of Finance}, 74.1(1):449--491.

\bibitem[Clark and McCracken, 2014]{clark2014tests}
Clark, T.~E. and McCracken, M.~W. (2014).
\newblock Tests of equal forecast accuracy for overlapping models.
\newblock {\em Journal of Applied Econometrics}, 29(3):415--430.

\bibitem[Clark and West, 2007]{clark2007approximately}
Clark, T.~E. and West, K.~D. (2007).
\newblock Approximately normal tests for equal predictive accuracy in nested
  models.
\newblock {\em Journal of econometrics}, 138(1):291--311.

\bibitem[Corsi, 2009]{corsi2009simple}
Corsi, F. (2009).
\newblock A simple approximate long-memory model of realized volatility.
\newblock {\em Journal of Financial Econometrics}, 7(2):174--196.

\bibitem[Cremers et~al., 2015]{cremers2015aggregate}
Cremers, M., Halling, M., and Weinbaum, D. (2015).
\newblock Aggregate jump and volatility risk in the cross-section of stock
  returns.
\newblock {\em The Journal of Finance}, 70(2):577--614.

\bibitem[Diebold and Mariano, 1995]{diebold2002comparing}
Diebold, F.~X. and Mariano, R.~S. (1995).
\newblock Comparing predictive accuracy.
\newblock {\em Journal of Business \& economic statistics}, 20(1):134--144.

\bibitem[Diebold and Y{\i}lmaz, 2014]{diebold2014network}
Diebold, F.~X. and Y{\i}lmaz, K. (2014).
\newblock On the network topology of variance decompositions: Measuring the
  connectedness of financial firms.
\newblock {\em Journal of Econometrics}, 182(1):119--134.

\bibitem[Doshi et~al., 2019]{doshi2019leverage}
Doshi, H., Jacobs, K., Kumar, P., and Rabinovitch, R. (2019).
\newblock Leverage and the cross-section of equity returns.
\newblock {\em The Journal of Finance}, 74(3):1431--1471.

\bibitem[Eugene and French, 1992]{eugene1992cross}
Eugene, F. and French, K. (1992).
\newblock The cross-section of expected stock returns.
\newblock {\em Journal of Finance}, 47(2):427--465.

\bibitem[Fan and Li, 2001]{fan2001variable}
Fan, J. and Li, R. (2001).
\newblock Variable selection via nonconcave penalized likelihood and its oracle
  properties.
\newblock {\em Journal of the American statistical Association},
  96(456):1348--1360.

\bibitem[Fan et~al., 2015]{fan2015power}
Fan, J., Liao, Y., and Yao, J. (2015).
\newblock Power enhancement in high-dimensional cross-sectional tests.
\newblock {\em Econometrica}, 83(4):1497--1541.

\bibitem[Gabaix, 2011]{gabaix2011granular}
Gabaix, X. (2011).
\newblock The granular origins of aggregate fluctuations.
\newblock {\em Econometrica}, 79(3):733--772.

\bibitem[Galvao, 2013]{galvao2013changes}
Galvao, A.~B. (2013).
\newblock Changes in predictive ability with mixed frequency data.
\newblock {\em International Journal of Forecasting}, 29(3):395--410.

\bibitem[Giacomini and White, 2006]{giacomini2006tests}
Giacomini, R. and White, H. (2006).
\newblock Tests of conditional predictive ability.
\newblock {\em Econometrica}, 74(6):1545--1578.

\bibitem[Giannone et~al., 2017]{giannone2017economic}
Giannone, D., Lenza, M., and Primiceri, G.~E. (2017).
\newblock Economic predictions with big data: The illusion of sparsity.

\bibitem[Hansen and Lunde, 2005]{hansen2005forecast}
Hansen, P.~R. and Lunde, A. (2005).
\newblock A forecast comparison of volatility models: does anything beat a
  garch (1, 1)?
\newblock {\em Journal of applied econometrics}, 20(7):873--889.

\bibitem[Hansen and Lunde, 2014]{hansen2014estimating}
Hansen, P.~R. and Lunde, A. (2014).
\newblock Estimating the persistence and the autocorrelation function of a time
  series that is measured with error.
\newblock {\em Econometric Theory}, 30(1):60--93.

\bibitem[Hirano and Wright, 2017]{hirano2017forecasting}
Hirano, K. and Wright, J.~H. (2017).
\newblock Forecasting with model uncertainty: Representations and risk
  reduction.
\newblock {\em Econometrica}, 85(2):617--643.

\bibitem[Hou, 2007]{hou2007industry}
Hou, K. (2007).
\newblock Industry information diffusion and the lead-lag effect in stock
  returns.
\newblock {\em The review of financial studies}, 20(4):1113--1138.

\bibitem[Inoue et~al., 2017]{inoue2017rolling}
Inoue, A., Jin, L., and Rossi, B. (2017).
\newblock Rolling window selection for out-of-sample forecasting with
  time-varying parameters.
\newblock {\em Journal of Econometrics}, 196(1):55--67.

\bibitem[Jurado et~al., 2015]{jurado2015measuring}
Jurado, K., Ludvigson, S.~C., and Ng, S. (2015).
\newblock Measuring uncertainty.
\newblock {\em American Economic Review}, 105(3):1177--1216.

\bibitem[Kapetanios et~al., 2014]{kapetanios2014nonlinear}
Kapetanios, G., Mitchell, J., and Shin, Y. (2014).
\newblock A nonlinear panel data model of cross-sectional dependence.
\newblock {\em Journal of Econometrics}, 179(2):134--157.

\bibitem[Kapetanios and Zikes, 2018]{kapetanios2018time}
Kapetanios, G. and Zikes, F. (2018).
\newblock Time-varying lasso.
\newblock {\em Economics Letters}, 169:1--6.

\bibitem[Klau et~al., 2018]{klau2018priority}
Klau, S., Jurinovic, V., Hornung, R., Herold, T., and Boulesteix, A.-L. (2018).
\newblock Priority-lasso: a simple hierarchical approach to the prediction of
  clinical outcome using multi-omics data.
\newblock {\em BMC bioinformatics}, 19(1):322.

\bibitem[Knight and Fu, 2000]{knight2000asymptotics}
Knight, K. and Fu, W. (2000).
\newblock Asymptotics for lasso-type estimators.
\newblock {\em Annals of statistics}, pages 1356--1378.

\bibitem[Li and Patton, 2018]{li2018asymptotic}
Li, J. and Patton, A.~J. (2018).
\newblock Asymptotic inference about predictive accuracy using high frequency
  data.
\newblock {\em Journal of Econometrics}, 203(2):223--240.

\bibitem[Li et~al., 2017a]{li2017jump}
Li, J., Todorov, V., and Tauchen, G. (2017a).
\newblock Jump regressions.
\newblock {\em Econometrica}, 85(1):173--195.

\bibitem[Li et~al., 2017b]{li2017Robust}
Li, J., Todorov, V., and Tauchen, G. (2017b).
\newblock Robust jump regressions.
\newblock {\em Journal of the American Statistical Association},
  112(517):332--341.

\bibitem[Liu et~al., 2020]{liu2020forecasting}
Liu, L., Moon, H.~R., and Schorfheide, F. (2020).
\newblock Forecasting with dynamic panel data models.
\newblock {\em Econometrica}, 88(1):171--201.

\bibitem[Liu et~al., 2015]{liu2015does}
Liu, L.~Y., Patton, A.~J., and Sheppard, K. (2015).
\newblock Does anything beat 5-minute rv? a comparison of realized measures
  across multiple asset classes.
\newblock {\em Journal of Econometrics}, 187(1):293--311.

\bibitem[Maddala, 1971]{maddala1971use}
Maddala, G.~S. (1971).
\newblock The use of variance components models in pooling cross section and
  time series data.
\newblock {\em Econometrica: Journal of the Econometric Society}, pages
  341--358.

\bibitem[McCracken, 2007]{mccracken2007asymptotics}
McCracken, M.~W. (2007).
\newblock Asymptotics for out of sample tests of granger causality.
\newblock {\em Journal of econometrics}, 140(2):719--752.

\bibitem[McGee and Olmo, 2020]{mcgee2020optimal}
McGee, R. and Olmo, J. (2020).
\newblock Optimal characteristic portfolios.
\newblock {\em Available at SSRN}.

\bibitem[Mundlak, 1978]{mundlak1978pooling}
Mundlak, Y. (1978).
\newblock On the pooling of time series and cross section data.
\newblock {\em Econometrica: journal of the Econometric Society}, pages 69--85.

\bibitem[Patton, 2011]{patton2011data}
Patton, A.~J. (2011).
\newblock Data-based ranking of realised volatility estimators.
\newblock {\em Journal of Econometrics}, 161(2):284--303.

\bibitem[Phillips, 1996]{phillips1996econometric}
Phillips, P.~C. (1996).
\newblock Econometric model determination.
\newblock {\em Econometrica: Journal of the Econometric Society}, pages
  763--812.

\bibitem[Tibshirani, 1996]{tibshirani1996regression}
Tibshirani, R. (1996).
\newblock Regression shrinkage and selection via the lasso.
\newblock {\em Journal of the Royal Statistical Society: Series B
  (Methodological)}, 58(1):267--288.

\bibitem[Wallace and Hussain, 1969]{wallace1969use}
Wallace, T.~D. and Hussain, A. (1969).
\newblock The use of error components models in combining cross section with
  time series data.
\newblock {\em Econometrica: Journal of the Econometric Society}, pages 55--72.

\bibitem[West, 1996]{west1996asymptotic}
West, K.~D. (1996).
\newblock Asymptotic inference about predictive ability.
\newblock {\em Econometrica: Journal of the Econometric Society}, pages
  1067--1084.

\bibitem[West, 2006]{west2006forecast}
West, K.~D. (2006).
\newblock Forecast evaluation.
\newblock {\em Handbook of economic forecasting}, 1:99--134.

\bibitem[White, 2000]{white2000reality}
White, H. (2000).
\newblock A reality check for data snooping.
\newblock {\em Econometrica}, 68(5):1097--1126.

\bibitem[Yao et~al., 2019]{yao2019novel}
Yao, X., Izzeldin, M., and Li, Z. (2019).
\newblock A novel cluster har-type model for forecasting realized volatility.
\newblock {\em Available at SSRN 3342090}.

\bibitem[Zou, 2006]{zou2006adaptive}
Zou, H. (2006).
\newblock The adaptive lasso and its oracle properties.
\newblock {\em Journal of the American statistical association},
  101(476):1418--1429.

\bibitem[Zou and Hastie, 2005]{zou2005regularization}
Zou, H. and Hastie, T. (2005).
\newblock Regularization and variable selection via the elastic net.
\newblock {\em Journal of the royal statistical society: series B (statistical
  methodology)}, 67(2):301--320.

\end{thebibliography}

\newpage

\end{document}